\documentclass{article}

\usepackage[final, nonatbib]{nips_2016}

\usepackage[utf8]{inputenc} 
\usepackage[T1]{fontenc}    
\usepackage{url}            
\usepackage{booktabs}       
\usepackage{amsfonts}       
\usepackage{nicefrac}       
\usepackage{microtype}      
\usepackage{tikz} 
\usepackage{xspace}
\usepackage{amsmath}
\usepackage{amssymb} 
\usepackage{amsfonts}
\usepackage{array}
\usepackage{verbatim}
\usepackage{wrapfig}
\usepackage{setspace}
\usetikzlibrary{shapes,arrows,matrix,positioning}

\newcommand{\x}{\mathbf{x}}
\newcommand{\nmax}{N}
\newcommand{\n}{n}
\newcommand{\z}{\mathbf{z}}

\newcommand{\zpres}{z_\textrm{pres}}

\newcommand{\zpresNN}{z_\textrm{pres}^{1:N}}
\newcommand{\zcode}{\mathbf{z}_\textrm{code}}
\newcommand{\zwhere}{\mathbf{z}_\textrm{where}}
\newcommand{\zwhereNN}{\mathbf{z}_\textrm{where}^{1:N}}
\newcommand{\zwhat}{\mathbf{z}_\textrm{what}}
\newcommand{\zwhatNN}{\mathbf{z}_\textrm{what}^{1:N}}
\newcommand{\xatt}{\mathbf{x}_\textrm{att}}
\newcommand{\yatt}{\mathbf{y}_\textrm{att}}
\newcommand{\y}{\mathbf{y}}
\newcommand{\h}{\mathbf{h}}
\newcommand{\paramP}{\theta}
\newcommand{\paramQ}{\phi}

\newcommand{\zsetn}{\mathbf{z}}

\newcommand{\zpresvec}{\mathbf{z_\textrm{pres}}}

\newcommand{\partialAt}[2]{\frac{\partial #1}{\partial #2}}
\newcommand{\expectationE}[2]{\mathbb{E}_{#2}  \left[ #1 \right]}
\newcommand{\diffd}{\textrm{d}}
\newcommand{\ls}{\left [ }
\newcommand{\rs}{\right ] }

\newcommand{\KL}{\textrm{KL}}

\newcommand{\MDL}{AIR\xspace}
\newcommand{\DMDL}{DAIR\xspace}
\newcommand{\Model}{Attend-Infer-Repeat\xspace}

\title{Attend, Infer, Repeat:\\Fast Scene Understanding with Generative Models}

\author{
  \textbf{S. M. Ali Eslami, Nicolas Heess, Theophane Weber, Yuval Tassa, }\\
  \textbf{David Szepesvari, Koray Kavukcuoglu, Geoffrey E. Hinton}\\
  \texttt{\{aeslami,heess,theophane,tassa,dsz,korayk,geoffhinton\}@google.com}\\
  Google DeepMind, London, UK
}

\begin{document}

\maketitle

\begin{abstract} 
We present a framework for efficient inference in structured image models that explicitly reason about objects. We achieve this by performing probabilistic inference using a recurrent neural network that attends to scene elements and processes them one at a time. Crucially, the model itself learns to choose the appropriate number of inference steps. We use this scheme to learn to perform inference in partially specified 2D models (variable-sized variational auto-encoders) and fully specified 3D models (probabilistic renderers). We show that such models learn to identify multiple objects -- counting, locating and classifying the elements of a scene -- without any supervision, e.g.,\ decomposing 3D images with various numbers of objects in a single forward pass of a neural network at unprecedented speed. We further show that the networks produce accurate inferences when compared to supervised counterparts, and that their structure leads to improved generalization. 
\end{abstract}

\section{Introduction}


The human percept of a visual scene is highly structured. Scenes naturally decompose into \emph{objects} that are arranged in space, have visual and physical properties, and are in functional relationships with each other. Artificial systems that interpret images in this way are desirable, as accurate detection of objects and inference of their attributes is thought to be fundamental for many problems of interest. Consider a robot whose task is to clear a table after dinner. To plan its actions it will need to determine which objects are present, what classes they belong to and where each one is located on the table.

The notion of using structured models for image understanding has a long history (e.g.,\ `vision as inverse graphics' \cite{grenander1976}), however in practice it has been difficult to define models that are: (a) expressive enough to capture the complexity of natural scenes, and (b) amenable to tractable inference. Meanwhile, advances in deep learning have shown how neural networks can be used to make sophisticated predictions from images using little interpretable structure (e.g.,\ \cite{krizhevsky2012}). Here we explore the intersection of structured probabilistic models and deep networks. Prior work on deep generative methods (e.g.,\ VAEs \cite{kingma2013}) have been mostly unstructured, therefore despite producing impressive samples and likelihood scores their representations have lacked interpretable meaning. On the other hand, structured generative methods have largely been incompatible with deep learning, and therefore inference has been hard and slow (e.g.,\ via MCMC).

Our proposed framework achieves scene interpretation via learned, amortized inference, and it imposes structure on its representation through appropriate partly- or fully-specified generative models, rather than supervision from labels. It is important to stress that by training generative models, the aim is not primarily to obtain good reconstructions, but to produce good representations, in other words to understand scenes. We show experimentally that by incorporating the right kinds of structures, our models produce representations that are more useful for downstream tasks than those produced by VAEs or state-of-the-art generative models such as DRAW \cite{gregor2015}.

The proposed framework crucially allows for reasoning about the complexity of a given scene (the  dimensionality of its latent space). We demonstrate that via an Occam's razor type effect, this makes it possible to discover the underlying causes of a dataset of images in an unsupervised manner. For instance, the model structure will enforce that a scene is formed by a variable number of entities that appear in different locations, but the process of learning will identify what these scene elements look like and where they appear in any given image. The framework also combines high-dimensional distributed representations with directly interpretable latent variables (e.g.,\ affine pose). This combination makes it easier to avoid the pitfalls of models that are too unconstrained (leading to data-hungry learning) or too rigid (leading to failure via mis-specification). 

The main contributions of the paper are as follows. First, in Sec.~\ref{approach} we formalize a scheme for efficient variational inference in latent spaces of variable dimensionality. The key idea is to treat inference as an \emph{iterative} process, implemented as a recurrent neural network that attends to one object at a time, and learns to use an \emph{appropriate number} of inference steps for each image. 
We call the proposed framework \textit{\Model} (\MDL). End-to-end learning is enabled by recent advances in amortized variational inference, e.g.,\ combining gradient based optimization for continuous latent variables with black-box optimization for discrete ones. Second, in Sec.~\ref{experiments} we show that \MDL allows for learning of generative models that decompose multi-object scenes into their underlying causes, e.g.,\ the constituent objects, in an unsupervised manner. We demonstrate these capabilities on MNIST digits (Sec.~\ref{mnist}), overlapping sprites and Omniglot glyphs (appendices \ref{appendix:sprites} and \ref{appendix:omniglot}). We show that model structure can provide an important inductive bias that is not easily learned otherwise, leading to improved generalization. Finally, in Sec.~\ref{3d} we demonstrate how our inference framework can be used to perform inference for a 3D rendering engine with unprecedented speed, recovering the counts, identities and 3D poses of complex objects in scenes with significant occlusion in a single forward pass of a neural network, providing a scalable approach to `vision as inverse graphics'.

\section{Approach} 
\label{approach}

In this paper we take a Bayesian perspective of scene interpretation, namely that of treating this task as inference in a generative model. Thus given an image $\x$ and a model $p^x_\paramP(\x|\z) p^z_\paramP(\z)$ parameterized by $\paramP$ we wish to recover the underlying scene description $\z$ by computing the posterior $p(\z|\x) = p^x_\paramP(\x|\z)p^z_\paramP(\z) / p(\x)$. In this view, the prior $p^z_\paramP(\z)$ captures our assumptions about the underlying scene, and the likelihood $p^x_\paramP(\x|\z)$ is our model of how a scene description is rendered to form an image. Both can take various forms depending on the problem at hand and we will describe particular instances in Sec.~\ref{experiments}. Together, they define the language that we use to describe a scene.

Many real-world scenes naturally decompose into objects. We therefore make the modeling assumption that the scene description is structured into groups of variables $\z^i$, where each group describes the attributes of one of the objects in the scene, e.g.,\ its type, appearance, and pose. Since the number of objects will vary from scene to scene, we assume models of the following form:
\begin{align}
p_\paramP(\x) = \sum_{\n=1}^{N} p_N(\n) \! \int \! p^z_\paramP(\zsetn | n) p^x_\paramP(\x | \zsetn) \diffd \zsetn. \label{eq:generalModel}
\end{align}
This can be interpreted as follows. We first sample the number of objects $n$ from a suitable prior (for instance a Binomial distribution) with maximum value $N$. The latent, variable length, scene descriptor $\zsetn=(\z^1,\z^2,\ldots,\z^n)$ is then sampled from a scene model $\zsetn \sim p^z_\paramP( \cdot| n)$. Finally, we render the image according to $\x \sim p^x_\paramP(\cdot | \zsetn)$. Since the indexing of objects is arbitrary, $p^z_\paramP(\cdot)$ is exchangeable and $p^x_\paramP(\x|\cdot)$ is permutation invariant, and therefore the posterior over $\z$ is exchangeable.

The prior and likelihood terms can take different forms. We consider two scenarios: For 2D scenes (Sec.~\ref{mnist}), each object is characterized in terms of a learned distributed continuous representation for its shape, and a continuous 3-dimensional variable for its pose (position and scale). For 3D scenes (Sec.~\ref{3d}), objects are defined in terms of a categorical variable that characterizes their identity, e.g.,\ sphere, cube or cylinder, as well as their positions and rotations. We refer to the two kinds of variables for each object $i$ in both scenarios as $\zwhat^i$ and $\zwhere^i$ respectively, bearing in mind that their meaning (e.g.,\ position and scale in pixel space vs.\ position and orientation in 3D space) and their data type (continuous  vs. discrete) will vary. We further assume that $\z^i$ are independent under the prior, i.e.,\ $p^z_\paramP(\zsetn|n) = \prod_{i=1}^n p^z_\paramP(\z^i)$, but non-independent priors, such as a distribution over hierarchical scene graphs (e.g.,\ \cite{zhu2006}), can also be accommodated. Furthermore, while the number of objects is bounded as per Eq.~\ref{eq:generalModel}, it is relatively straightforward to relax this assumption.  

\subsection{Inference}
\label{approach-inference}

\begin{figure}
\begin{centering}
\includegraphics[height=2.7cm]{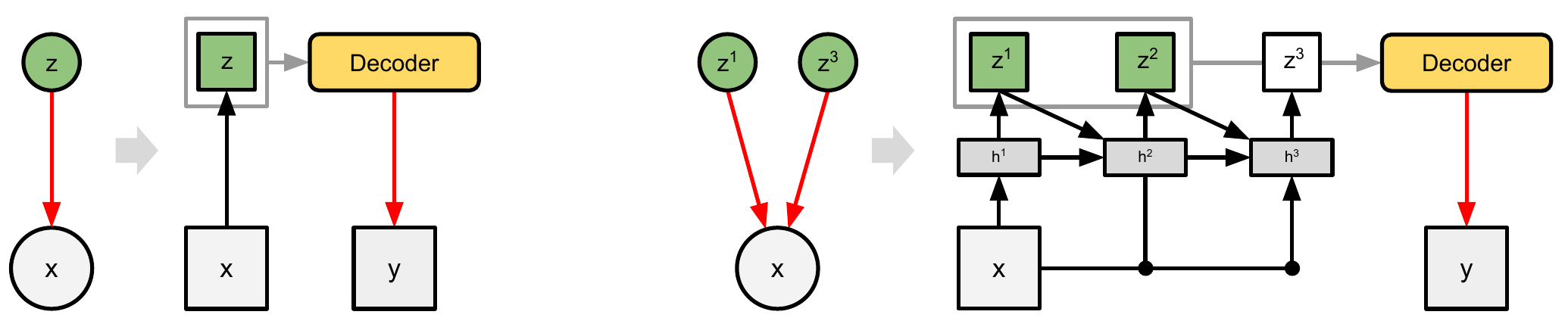}
\caption{\textit{Left:} A single random variable $z$ produces the observation $x$ (the image). The relationship between $z$ and $x$ is specified by a model. Inference is the task of computing likely values of $z$ given $x$. Using an auto-encoding architecture, the model (red arrow) and its inference network (black arrow) can be trained end-to-end via gradient descent. \textit{Right:} For most images of interest, multiple latent variables (e.g.,\ multiple objects) give rise to the image. We propose an iterative, variable-length inference network (black arrows) that attends to one object at a time, and train it jointly with its model. The result is fast, feed-forward, interpretable scene understanding trained without supervision.}
\label{models}
\end{centering}
\end{figure} 

Despite their natural appeal, inference for most models in the form of Eq.~\ref{eq:generalModel} is intractable. We therefore employ an amortized variational approximation to the true posterior by learning a distribution $q_\paramQ(\zsetn, \n | \x)$ parameterized by $\paramQ$ that minimizes the divergence $\KL \ls q_\paramQ(\zsetn, \n | \x) || p^z_\paramP(\zsetn, \n | \x) \rs$. While amortized variational approximations have recently been used successfully in a variety of works \cite{rezende2014,kingma2013,mnih2014b} the specific form of our model poses two additional difficulties. \textit{Trans-dimensionality:} As a challenging departure from classical latent space models, the size of the latent space $n$ (i.e.,\ the number of objects) is a random variable itself, which necessitates evaluating $p_N(n | \x) = \int p^z_\paramP(\zsetn, n |  x) \diffd \zsetn$, for all $n=1...N$. \textit{Symmetry:} There are strong symmetries that arise, for instance, from alternative assignments of objects appearing in an image $\x$ to latent variables $\z^i$.

We address these challenges by formulating inference as an iterative process implemented as a recurrent neural network, which infers the attributes of one object at a time. The network is run for $\nmax$ steps and in each step explains one object in the scene, conditioned on the image and on its knowledge of previously explained objects (see Fig.~\ref{models}). 

To simplify sequential reasoning about the number of objects, we parameterize $n$ as a variable length latent vector $\zpresvec$ using a unary code: for a given value $n$, $\zpresvec$ is the vector formed of $n$ ones followed by one zero. Note that the two representations are equivalent. The posterior takes the following form:
\begin{align}
 q_\paramQ(\z, \zpresvec|\x) = q_\paramQ(\zpres^{n+1}=0 | \z^{1:n}, \mathbf{x}) \prod_{i=1}^{n} & q_\paramQ(\z^i, \zpres^i =1 | \x, \z^{1:i-1}).
\end{align}

$q_\paramQ$ is implemented as a neural network that, in each step, outputs the parameters of the sampling distributions over the latent variables, e.g.,\ the mean and standard deviation of a Gaussian distribution for continuous variables. $\zpres$ can be understood as an interruption variable: at each time step, if the network outputs $\zpres=1$, it describes at least one more object and proceeds, but if it outputs $\zpres=0$, no more objects are described, and inference terminates for that particular datapoint.

Note that conditioning of $\z^i | \x, \z^{1:i-1}$ is critical to capture dependencies between the latent variables $\z^i$ in the posterior, e.g.,\ to avoid explaining the same object twice. The specifics of the networks that achieve this depend on the particularities of the models and we will describe them in detail in Sec.~\ref{experiments}. 

\subsection{Learning}
\label{approach-learning}

We can jointly optimize the parameters $\paramP$ of the model and $\paramQ$ of the inference network by maximizing the lower bound on the marginal likelihood of an image under the model: $\log p_\theta(\x) \geq \mathcal{L}(\paramP, \paramQ) = \expectationE{ \log \frac{ p_\theta(\x, \zsetn , n)}{q_\paramQ(\z, n, | \x) } }{q_\paramQ} \label{eq:VLB}$ with respect $\paramP$ and $\paramQ$. $\mathcal{L}$ is called the negative free energy. We provide an outline of how to construct an estimator of the gradient of this quantity below, for more details see \cite{schulman2015}.

Computing a Monte Carlo estimate of $\partialAt{}{\paramP} \mathcal{L}$ is relatively straightforward: given a sample from the approximate posterior $(\z, \zpresvec) \sim q_\paramQ(\cdot | \x)$ (i.e.,\ when the latent variables have been `filled in') we can readily compute $\partialAt{}{\paramP}\log p_\paramP(\x, \zsetn, n)$ provided $p$ is differentiable in $\paramP$.

Computing a Monte Carlo estimate of $\partialAt{}{\paramQ} \mathcal{L}$ is more involved. As discussed above, the RNN that implements $q_\paramQ$ produces the parameters of the sampling distributions for the scene variables $\zsetn$ and presence variables $\zpresvec$. For a time step $i$, denote with $\omega^i$ all the parameters of the sampling distributions of variables in $(\zpres^i,\z^i)$.
We parameterize the dependence of this distribution on $\z^{1:i-1}$ and $\x$ using a recurrent function $R_\paramQ (\cdot)$ implemented as a neural network such that $(\omega^i, \h^i) = R_\paramQ (\x, \h^{i-1})$ with hidden variables $\h$. The full gradient is obtained via chain rule: $\partial \mathcal{L} / \partial \paramQ  = \sum_{i} \partial \mathcal{L} / \partial \omega^i \times \partial \omega^i / \paramQ.$ Below we explain how to compute $\partial \mathcal{L} / \partial \omega^i$. We first rewrite our cost function as follows: $\mathcal{L}(\theta,\phi)=\expectationE{\ell(\theta,\phi,\zsetn,n)}{q_\paramQ}$ where $\ell(\theta,\phi,\zsetn,n)$ is defined as $\log \frac{ p_\theta(\x, \zsetn , n)}{q_\paramQ(\z, n, | \x) } $. Let $z^i$ be an arbitrary element of the vector $(\z^i,\zpres^i)$ of type \{what, where, pres\}. How to proceed depends on whether $z^i$ is continuous or discrete.

\paragraph{Continuous:} 
Suppose $z^i$ is a continuous variable. We use the path-wise estimator (also known as the `re-parameterization trick', e.g.,\ \cite{kingma2013,schulman2015}), which allows us to `back-propagate' through the random variable $z^i$. For many continuous variables (in fact, without loss of generality), $z^i$ can be sampled as $h(\xi,\omega^i)$, where $h$ is a deterministic transformation function, and $\xi$ a random variable from a fixed noise distribution $p(\xi)$ giving the gradient estimate: $\partialAt{\mathcal{L}}{\omega^i} \approx \partial \ell(\theta,\phi,\zsetn,n) / \partial z^i \times \partial h / \partial \omega^i$.

\paragraph{Discrete:}
For discrete scene variables (e.g.,\ $\zpres^i$) we cannot compute the gradient $\partial \mathcal{L} / \partial \omega^i_j $ by back-propagation. Instead we use the likelihood ratio estimator \cite{mnih2014b,schulman2015}. Given a posterior sample $(\z,n) \sim q_\paramQ(\cdot | \x)$ we can obtain a Monte Carlo estimate of the gradient: $\partial \mathcal{L} / \partial \omega^i \approx \partial \log q( z^i | \omega^i) / \partial \omega^i \: \ell(\theta,\phi,\z,n).$ In the raw form presented here this gradient estimate is likely to have high variance. We reduce its variance using appropriately structured neural baselines \cite{mnih2014b} that are functions of the image and the latent variables produced so far.

\section{Models and Experiments}
\label{experiments}
We first apply \MDL to a dataset of multiple MNIST digits, and show that it can reliably learn to detect and generate the constituent digits from scratch (Sec.~\ref{mnist}). We show that this provides advantages over state-of-the-art generative models such as DRAW \cite{gregor2015} in terms of computational effort, generalization to unseen datasets, and the usefulness of the inferred representations for downstream tasks. We also apply \MDL to a setting where a 3D renderer is specified in advance. We show that \MDL learns to use the renderer to infer the counts, identities and poses of multiple objects in synthetic and real table-top scenes with unprecedented speed (Sec.~\ref{3d} and appendix \ref{appendix:speed}).

\begin{figure}
\begin{center}
\includegraphics[height=2.7cm]{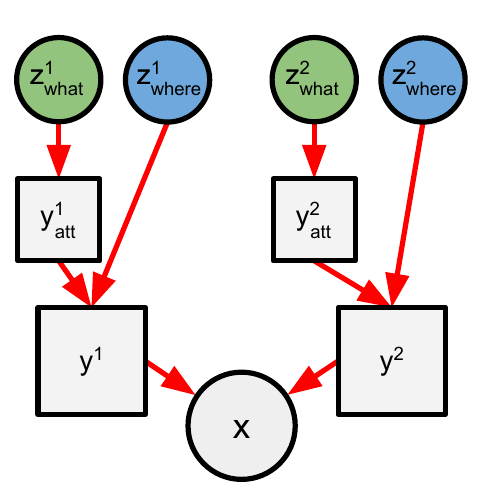}
\hfill
\includegraphics[height=2.7cm]{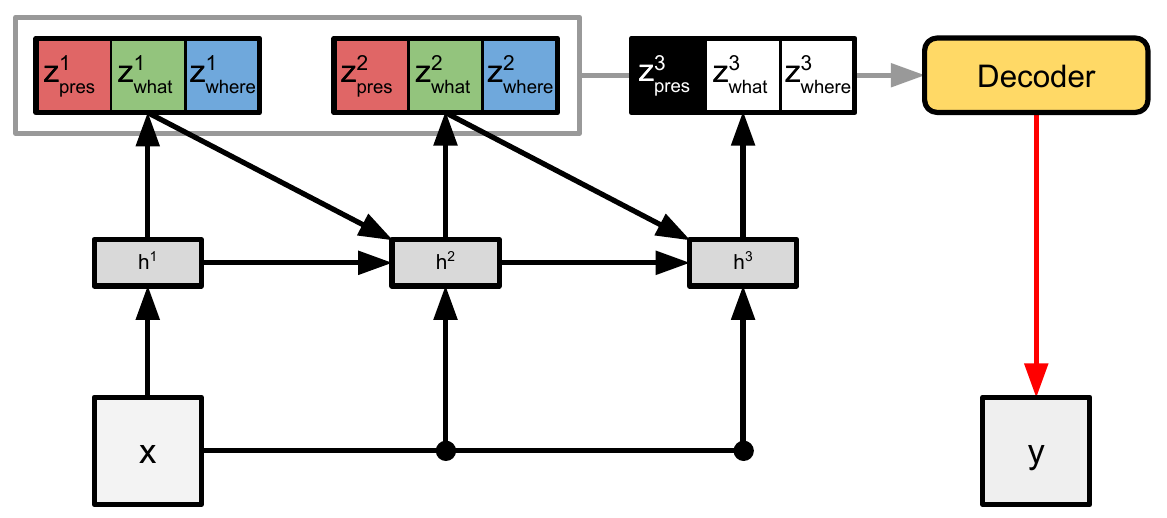}
\hfill
\includegraphics[height=2.7cm]{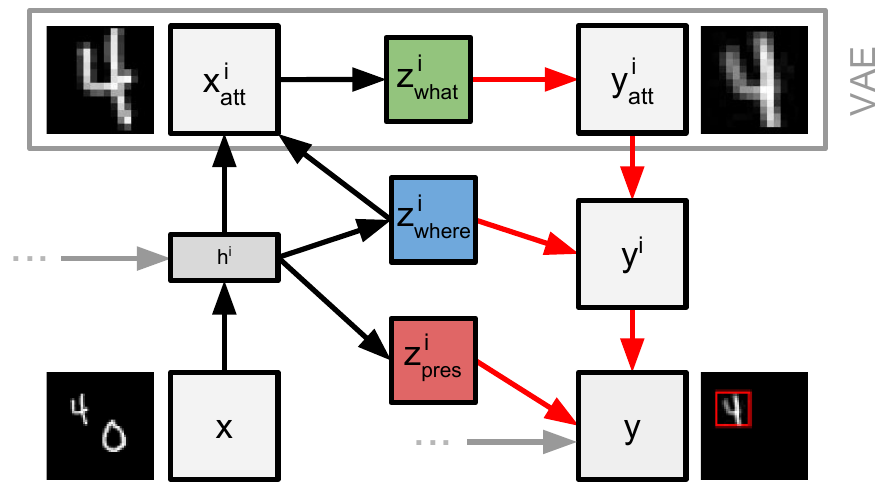}
\vspace{-0.2cm}
\caption{\textbf{\MDL in practice:} \textit{Left:} The assumed generative model. \textit{Middle:} \MDL inference for this model. The contents of the grey box are input to the decoder. \textit{Right:} Interaction between the inference and generation networks at every time-step. In our experiments the relationship between $\xatt^i$ and $\yatt^i$ is modeled by a VAE, however any generative model of patches could be used (even, e.g.,\ DRAW).}
\label{models-mdl-mnist}
\end{center}
\end{figure}

Details of the \MDL model and networks used in the 2D experiments are shown in Fig.~\ref{models-mdl-mnist}. The generative model (Fig.~\ref{models-mdl-mnist}, left) draws $n \sim \textrm{Geom}(\rho)$ digits $\{ \yatt^i \}$, scales and shifts them according to $\zwhere^i \sim \mathcal{N}(0, \Sigma)$ using spatial transformers, and sums the results $\{ y^i \}$ to form the image. Each digit is obtained by first sampling a latent code $\zwhat^i$ from the prior $\zwhat^i \sim \mathcal{N}(\mathbf{0},\mathbf{1})$ and propagating it through a decoder network. The learnable parameters of the generative model are the parameters of this decoder network. The \MDL inference network (Fig.~\ref{models-mdl-mnist}, middle) produces three sets of variables for each entity at every time-step: a 1-dimensional Bernoulli variable indicating the entity's presence, a $C$-dimensional distributed vector describing its class or appearance ($\zwhat^i$), and a 3-dimensional vector specifying the affine parameters of its position and scale ($\zwhere^i$). Fig.~\ref{models-mdl-mnist} (right) shows the interaction between the inference and generation networks at every time-step. The inferred pose is used to attend to a part of the image (using a spatial transformer) to produce $\xatt^i$, which is processed to produce the inferred code $\zcode^i$ and the reconstruction of the contents of the attention window $\yatt^i$. The same pose information is used by the generative model to transform $\yatt^i$ to obtain $\y^i$. This contribution is only added to the canvas $\y$ if $\zpres^i$ was inferred to be true.

For the dataset of MNIST digits, we also investigate the behavior of a variant, difference-\MDL (\DMDL), which employs a slightly different recurrent architecture for the inference network (see Fig.~\ref{models-dmdl-mnist} in appendix). As opposed to \MDL which computes $\z^i$ via $\h^i$ and $\x$, \DMDL reconstructs at every time step $i$ a partial reconstruction $\x^i$ of the data $\x$, which is set as the mean of the distribution $p_\paramP^x(\x|\z^1,\z^2,\ldots,\z^{i-1})$. We create an error canvas $\Delta \x^i = \x^i-\x$, and the \DMDL inference equation $R_\phi$ is then specified as $(\omega^i, \h^i) = R_\paramQ (\Delta \x^i, \h^{i-1})$.

\subsection{Multi-MNIST}
\label{mnist}

\begin{figure}
\vspace{-0.3cm}
\begin{center}
\begin{minipage}{0.6\linewidth}
\begin{tikzpicture}
\matrix [matrix of nodes, column sep=1mm, row sep=1mm, every node/.style={inner sep=0, outer sep=0, anchor=center}]
{
  \node[rotate=90]{\small{Data}}; & 
  \includegraphics[height=1cm, width=1cm]{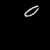} & 
  \includegraphics[height=1cm, width=1cm]{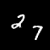} & 
  \includegraphics[height=1cm, width=1cm]{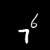} & 
  \includegraphics[height=1cm, width=1cm]{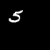} & 
  \includegraphics[height=1cm, width=1cm]{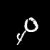} & 
  \includegraphics[height=1cm, width=1cm]{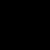} & 
  \includegraphics[height=1cm, width=1cm]{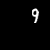} \\
  \vspace{1mm} \\
  \node[rotate=90]{\small{1k}}; & 
  \includegraphics[height=1cm, width=1cm]{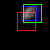} & 
  \includegraphics[height=1cm, width=1cm]{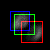} & 
  \includegraphics[height=1cm, width=1cm]{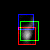} & 
  \includegraphics[height=1cm, width=1cm]{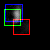} & 
  \includegraphics[height=1cm, width=1cm]{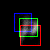} & 
  \includegraphics[height=1cm, width=1cm]{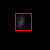} & 
  \includegraphics[height=1cm, width=1cm]{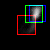} \\
  \node[rotate=90]{\small{10k}}; & 
  \includegraphics[height=1cm, width=1cm]{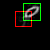} & 
  \includegraphics[height=1cm, width=1cm]{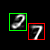} & 
  \includegraphics[height=1cm, width=1cm]{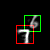} & 
  \includegraphics[height=1cm, width=1cm]{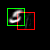} & 
  \includegraphics[height=1cm, width=1cm]{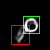} & 
  \includegraphics[height=1cm, width=1cm]{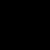} & 
  \includegraphics[height=1cm, width=1cm]{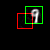} \\
  \node[rotate=90]{\small{200k}}; & 
  \includegraphics[height=1cm, width=1cm]{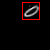} & 
  \includegraphics[height=1cm, width=1cm]{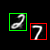} & 
  \includegraphics[height=1cm, width=1cm]{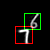} & 
  \includegraphics[height=1cm, width=1cm]{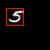} & 
  \includegraphics[height=1cm, width=1cm]{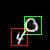} & 
  \includegraphics[height=1cm, width=1cm]{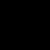} & 
  \includegraphics[height=1cm, width=1cm]{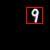} \\
};
\end{tikzpicture}
\end{minipage}
\hspace{0.1cm}
\begin{minipage}{0.29\linewidth}
\includegraphics[width=\linewidth]{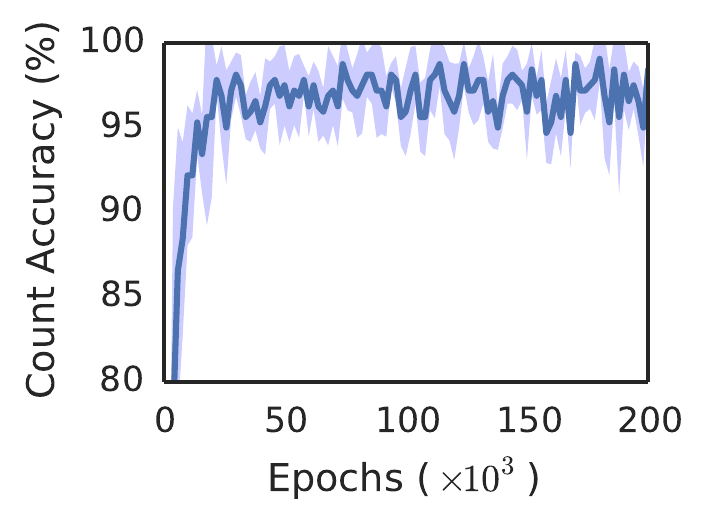}
\\
\begin{tikzpicture}
\matrix [matrix of nodes, column sep=1mm, row sep=1mm, every node/.style={inner sep=1mm, outer sep=0, anchor=center}]
{
  \includegraphics[width=1cm]{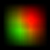} & 
  \includegraphics[width=1cm]{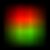} &
  \includegraphics[width=1cm]{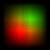} \\
  \draw[thick,-stealth] (0.1,0.1) -- (-0.1,-0.1); &
  \draw[thick,-stealth] (0.0,0.15) -- (-0.0,-0.15); &
  \draw[thick,-stealth] (-0.1,-0.1) -- (0.1,0.1); \\
};
\end{tikzpicture}
\end{minipage}
\vspace{-0.1cm}
\caption{\textbf{Multi-MNIST learning:} \textit{Left above:} Images from the dataset. \textit{Left below:} Reconstructions at different stages of training along with a visualization of the model's attention windows. The 1st, 2nd and 3rd time-steps are displayed using red, green and blue borders respectively. A video of this sequence is provided in the supplementary material. \textit{Above right:} Count accuracy over time. The model detects the counts of digits accurately, despite having never been provided supervision. \textit{Below right:} The learned scanning policy for 3 different runs of training (only differing in the random seed). We visualize empirical heatmaps of the attention windows' positions (red, and green for the first and second time-steps respectively). As expected, the policy is random. This suggests that the policy is spatial, as opposed to identity- or size-based.}
\vspace{-0.2cm}
\label{mnist-reconstructions-over-time}
\end{center}
\end{figure} 

We begin with a 50$\times$50 dataset of multi-MNIST digits. Each image contains zero, one or two non-overlapping random MNIST digits with equal probability. The desired goal is to train a network that produces sensible explanations for each of the images. We train \MDL with $N=3$ on 60,000 such images from scratch, i.e.,\ without a curriculum or any form of supervision by maximizing $\mathcal{L}$ with respect to the parameters of the inference network and the generative model. Upon completion of training we inspect the model's inferences (see Fig.~\ref{mnist-reconstructions-over-time}, left). We draw the reader's attention to the following observations. First, the model identifies the number of digits correctly, due to the opposing pressures of (a) wanting to explain the scene, and (b) the cost that arises from instantiating an object under the prior. This is indicated by the number of attention windows in each image; we also plot the accuracy of count inference over the course of training (Fig.~\ref{mnist-reconstructions-over-time}, above right). Second, it locates the digits accurately. Third, the recurrent network learns a suitable scanning policy to ensure that different time-steps account for different digits (Fig.~\ref{mnist-reconstructions-over-time}, below right). Note that we did not have to specify any such policy in advance, nor did we have to build in a constraint to prevent two time-steps from explaining the same part of the image. Finally, that the network learns to not use the second time-step when the image contains only a single digit, and to never use the third time-step (images contain a maximum of two digits). This allows for the inference network to stop upon encountering the first $\zpres^i$ equaling 0, leading to potential savings in computation during inference.

A video showing real-time inference using \MDL has been included in the supplementary material. We also perform experiments on Omniglot (\cite{lake2015}, appendix~\ref{appendix:omniglot}) to demonstrate \MDL's ability to parse glyphs into elements resembling `strokes', as well as a dataset of sprites where the scene's elements appear under significant overlap (appendix~\ref{appendix:sprites}). See appendices for details and results.

\begin{figure}[t!]
\begin{center}
\begin{minipage}{0.4\linewidth}
\begin{tikzpicture}
\matrix [matrix of nodes, column sep=1mm, row sep=1mm, every node/.style={inner sep=0, outer sep=0, anchor=center}]
{
  \node[rotate=90]{\small{Data}}; & 
  \includegraphics[height=1cm, width=1cm]{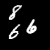} & 
  \includegraphics[height=1cm, width=1cm]{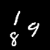} \\
  \\
  \node[rotate=90]{\small{\DMDL}}; & 
  \includegraphics[height=1cm, width=1cm]{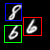} & 
  \includegraphics[height=1cm, width=1cm]{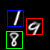} \\ 
  \\
  \\
  \\
};
\end{tikzpicture}
\begin{tikzpicture}
\matrix [matrix of nodes, column sep=1mm, row sep=1mm, every node/.style={inner sep=0, outer sep=0, anchor=center}]
{
  \node[rotate=90]{\small{Data}}; & 
  \includegraphics[height=1cm, width=1cm]{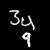} & 
  \includegraphics[height=1cm, width=1cm]{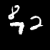} \\
  \\
  \node[rotate=90]{\small{DRAW}}; & 
  \includegraphics[height=1cm, width=1cm]{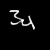} & 
  \includegraphics[height=1cm, width=1cm]{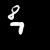} \\ 
  \\
  \\
  \\
};
\end{tikzpicture}
\end{minipage}
\begin{minipage}{0.58\linewidth}
  \centering
  \includegraphics[width=0.49\linewidth]{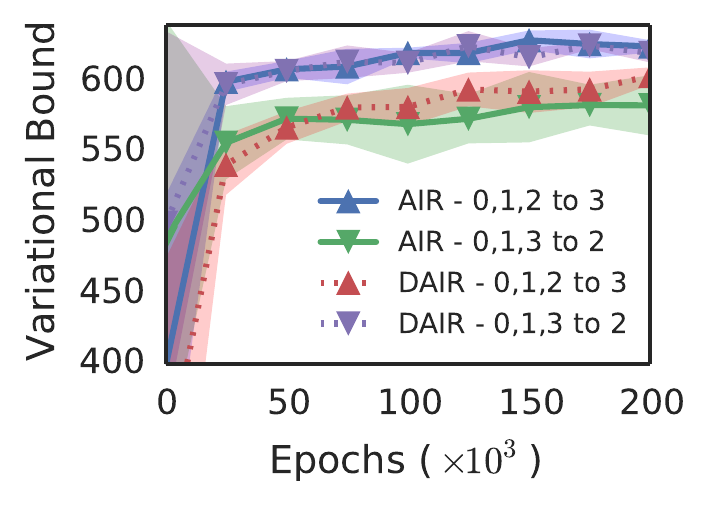}
  \includegraphics[width=0.49\linewidth]{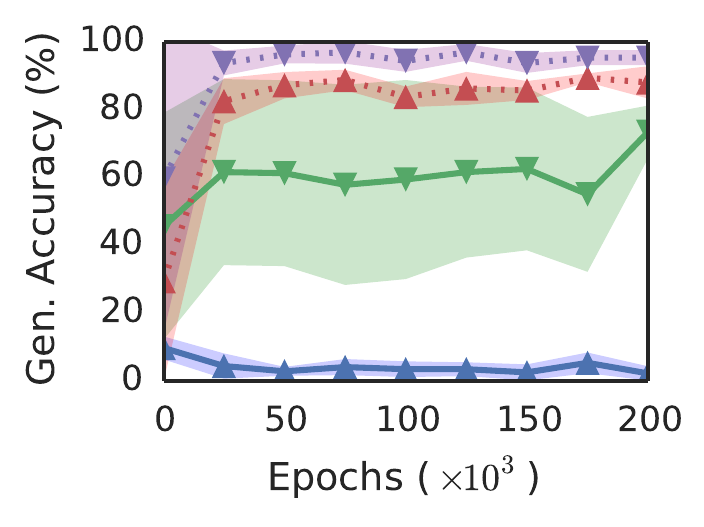}
\end{minipage}
\vspace{-0.2cm}
\caption{\textbf{Strong generalization:} \textit{Left:} Reconstructions of images with 3 digits made by \DMDL trained on 0, 1 or 2 digits, as well as a comparison with DRAW. \textit{Right:} Variational lower bound, and generalizing / interpolating count accuracy. \DMDL out-performs both DRAW and \MDL at this task.}
\vspace{-0.5cm}
\label{mnist-generalisation-curves} 
\end{center}
\end{figure} 

\subsubsection{Strong Generalization} Since the model learns the concept of a digit independently of the positions or numbers of times it appears in each image, one would hope that it would be able to generalize, e.g.,\ by demonstrating an understanding of scenes that have structural differences to training scenes. We probe this behavior with the following scenarios: (a) \textit{Extrapolation:} training on images each containing 0, 1 or 2 digits and then testing on images containing 3 digits, and (b) \textit{Interpolation:} training on images containing 0, 1 or 3 digits and testing on images containing 2 digits. The result of this experiment is shown in Fig.~\ref{mnist-generalisation-curves}. An \MDL model trained on up to 2 digits is effectively unable to infer the correct count when presented with an image of 3 digits. We believe this to be caused by the LSTM which learns during training never to expect more than 2 digits. \MDL's generalization performance is improved somewhat when considering the interpolation task. \DMDL by contrast generalizes well in both tasks (and finds interpolation to be slightly easier than extrapolation). A closely related baseline is the Deep Recurrent Attentive Writer (DRAW, \cite{gregor2015}), which like \MDL, generates data sequentially. However, DRAW has a fixed and large number of steps (40 in our experiments). As a consequence generative steps do not correspond to easily interpretable entities, complex scenes are drawn faster and simpler ones slower. We show DRAW's reconstructions in Fig.~\ref{mnist-generalisation-curves}. Interestingly, DRAW learns to ignore precisely one digit in the image. See appendix for further details of these experiments.

\subsubsection{Representational Power}

\begin{wrapfigure}{r}{0.5\textwidth}
\vspace{-1.5cm}
\begin{center}
\includegraphics[width=0.49\linewidth]{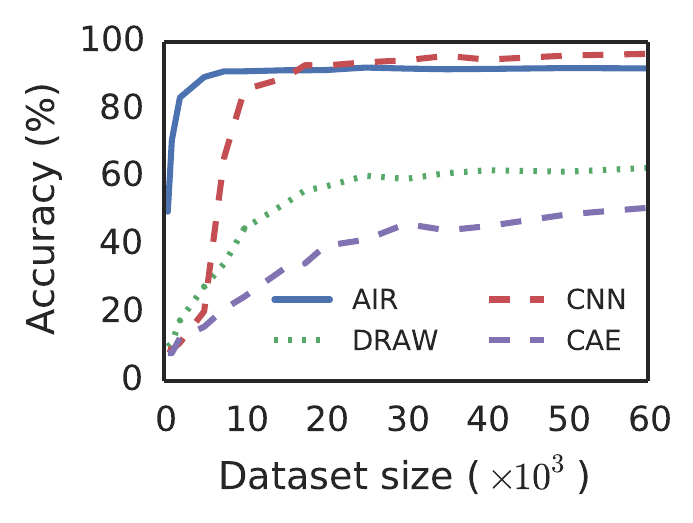}
\hfill
\includegraphics[width=0.49\linewidth]{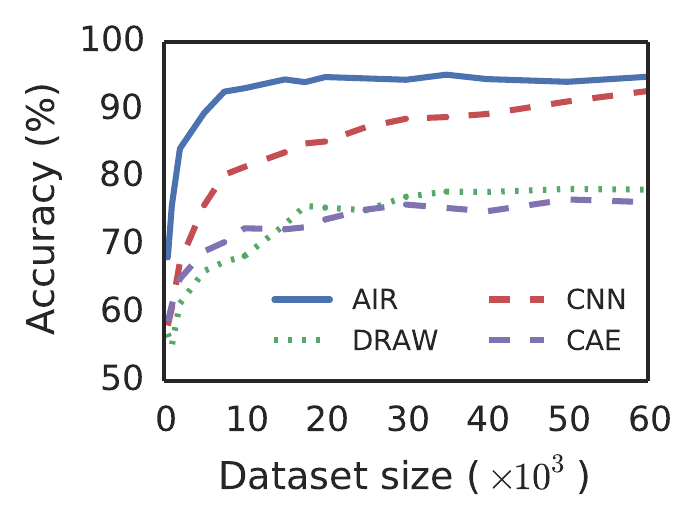}
\vspace{-0.7cm}
\caption{\textbf{Representational power:} \MDL achieves high accuracy using only a fraction of the labeled data. \textit{Left:} summing two digits. \textit{Right:} detecting if they appear in increasing order. Despite producing comparable reconstructions, CAE and DRAW inferences are less interpretable than \MDL's and therefore lead to poorer downstream performance.}
\label{mnist-air-vs-cnn}
\end{center}
\end{wrapfigure} 

A second motivation for the use of structured models is that their inferences about a scene provides useful representations for downstream tasks. We examine this ability by first training an \MDL model on 0, 1 or 2 digits and then produce inferences for a separate collection of images that contains precisely 2 digits. We split this data into training and test and consider two tasks: (a) predicting the sum of the two digits (as was done in \cite{ba2015}), and (b) determining if the digits appear in an ascending order. We compare with a CNN trained from the raw pixels, as well as interpretations produced by a convolutional autoencoder (CAE) and DRAW (Fig.~\ref{mnist-air-vs-cnn}). We optimize each model's hyper-parameters (e.g.\, depth and size) for maximal performance. \MDL achieves high accuracy even when data is scarce, indicating the power of its disentangled, structured representation. See appendix for further details.

\subsection{3D Scenes}
\label{3d}

The experiments above demonstrate learning of inference \textit{and} generative networks in models where we impose structure in the form of a variable-sized representation and spatial attention mechanisms. We now consider an additional way of imparting knowledge to the system: we specify the generative model via a 3D renderer, i.e.,\ we completely specify how any scene representation is transformed to produce the pixels in an image. Therefore the task is to learn to infer the counts, identities and poses of several objects, given different images containing these objects and an implementation of a 3D renderer from which we can draw new samples. This formulation of computer vision is often called `vision as inverse graphics' (see e.g.,\ \cite{grenander1976,loper2014,jampani2015}).

The primary challenge in this view of computer vision is that of inference. While it is relatively easy to specify high-quality models in the form of probabilistic renderers, posterior inference is either extremely expensive or prone to getting stuck in local minima (e.g.,\ via optimization or MCMC). In addition, probabilistic renderers (and in particular renderers) typically are not capable of providing gradients with respect to their inputs, and 3D scene representations often involve discrete variables, e.g.,\ mesh identities. We address these challenges by using finite-differencing to obtain a gradient through the renderer, using the score function estimator to get gradients with respect to discrete variables, and using \MDL inference to handle correlated posteriors and variable-length representations.

\begin{figure}
\begin{center}
\begin{tikzpicture}
\matrix [matrix of nodes, column sep=1mm, row sep=1mm, every node/.style={inner sep=0, outer sep=0, anchor=center}]
{
  \node[rotate=90]{\small{\textit{(a)} Data}}; & 
  \includegraphics[height=1cm, width=1cm]{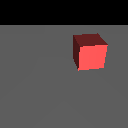} & 
  \includegraphics[height=1cm, width=1cm]{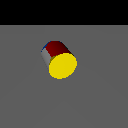} & 
  \includegraphics[height=1cm, width=1cm]{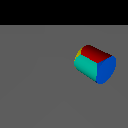} & 
  \includegraphics[height=1cm, width=1cm]{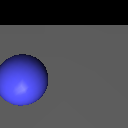} \\
  \vspace{1mm} \\
  \node[rotate=90]{\small{\textit{(b)} \MDL}}; & 
  \includegraphics[height=1cm, width=1cm]{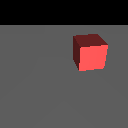} & 
  \includegraphics[height=1cm, width=1cm]{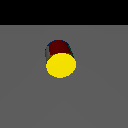} &  
  \includegraphics[height=1cm, width=1cm]{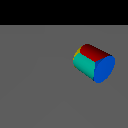} & 
  \includegraphics[height=1cm, width=1cm]{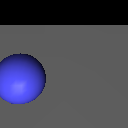} \\
  \node[rotate=90]{\small{\textit{(c)} Sup.}}; & 
  \includegraphics[height=1cm, width=1cm]{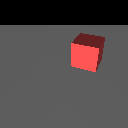} & 
  \includegraphics[height=1cm, width=1cm]{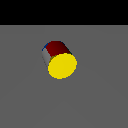} & 
  \includegraphics[height=1cm, width=1cm]{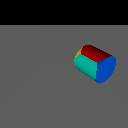} & 
  \includegraphics[height=1cm, width=1cm]{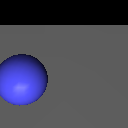} \\
  \node[rotate=90]{\small{\textit{(d)} Opt.}}; & 
  \includegraphics[height=1cm, width=1cm]{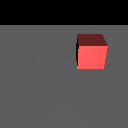} & 
  \includegraphics[height=1cm, width=1cm]{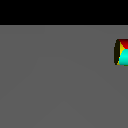} & 
  \includegraphics[height=1cm, width=1cm]{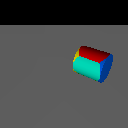} & 
  \includegraphics[height=1cm, width=1cm]{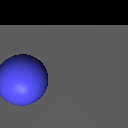}\\
};
\end{tikzpicture}
\begin{tikzpicture}
\matrix [matrix of nodes, column sep=1mm, row sep=1mm, every node/.style={inner sep=0, outer sep=0, anchor=center}]
{
  \node[rotate=90]{\small{\textit{(e)} Data}}; &
  \includegraphics[height=1cm, width=1cm]{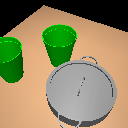} & 
  \includegraphics[height=1cm, width=1cm]{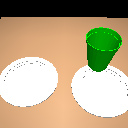} & 
  \includegraphics[height=1cm, width=1cm]{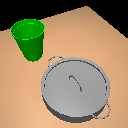} & 
  \includegraphics[height=1cm, width=1cm]{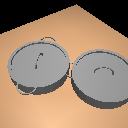} & 
  \includegraphics[height=1cm, width=1cm]{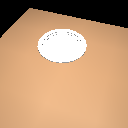} & 
  \includegraphics[height=1cm, width=1cm]{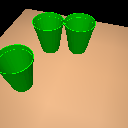} & 
  \includegraphics[height=1cm, width=1cm]{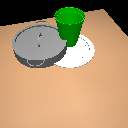} \\
  \node[rotate=90]{\small{\textit{(f)} \MDL}}; & 
  \includegraphics[height=1cm, width=1cm]{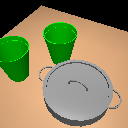} & 
  \includegraphics[height=1cm, width=1cm]{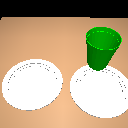} & 
  \includegraphics[height=1cm, width=1cm]{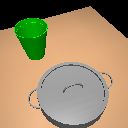} & 
  \includegraphics[height=1cm, width=1cm]{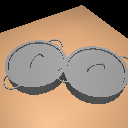} & 
  \includegraphics[height=1cm, width=1cm]{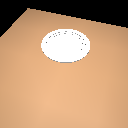} & 
  \includegraphics[height=1cm, width=1cm]{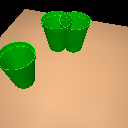} & 
  \includegraphics[height=1cm, width=1cm]{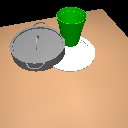} \\
  \\
  \node[rotate=90]{\small{\textit{(g)} Real}}; &
  \includegraphics[height=1cm, width=1cm]{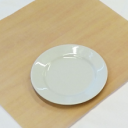} & 
  \includegraphics[height=1cm, width=1cm]{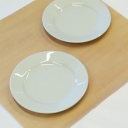} & 
  \includegraphics[height=1cm, width=1cm]{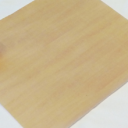} & 
  \includegraphics[height=1cm, width=1cm]{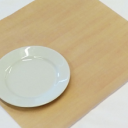} & 
  \includegraphics[height=1cm, width=1cm]{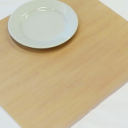} & 
  \includegraphics[height=1cm, width=1cm]{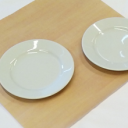} & 
  \includegraphics[height=1cm, width=1cm]{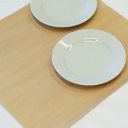} \\
  \node[rotate=90]{\small{\textit{(h)} \MDL}}; & 
  \includegraphics[height=1cm, width=1cm]{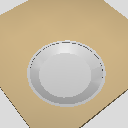} & 
  \includegraphics[height=1cm, width=1cm]{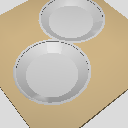} & 
  \includegraphics[height=1cm, width=1cm]{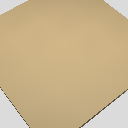} & 
  \includegraphics[height=1cm, width=1cm]{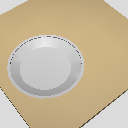} & 
  \includegraphics[height=1cm, width=1cm]{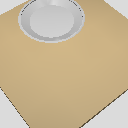} & 
  \includegraphics[height=1cm, width=1cm]{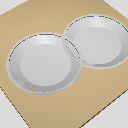} & 
  \includegraphics[height=1cm, width=1cm]{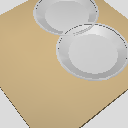} \\
};
\end{tikzpicture}
\caption{\textbf{3D objects:} \textit{Left:} The task is to infer the identity and pose of a single 3D object. (a) Images from the dataset. (b) Unsupervised \MDL reconstructions. (c) Supervised reconstructions. Note poor performance on cubes due to their symmetry. (d) Reconstructions after direct gradient descent. This approach is less stable and much more susceptible to local minima. \textit{Right:} \MDL can learn to recover the counts, identities and poses of multiple objects in a 3D table-top scene. (e,g) Generated and real images. (f,h) \MDL produces fast and accurate inferences which we visualize using the renderer.}
\vspace{-0.8cm}
\label{3d-single}
\end{center}
\end{figure}

We demonstrate the capabilities of this approach by first considering scenes consisting of only one of three objects: a red cube, a blue sphere, and a textured cylinder (see Fig.~\ref{3d-single}a). Since the scenes only consist of single objects, the task is only to infer the identity (cube, sphere, cylinder) and pose (position and rotation) of the object present in the image. We train a single-step ($N=1$) \MDL inference network for this task. The network is only provided with unlabeled images and is trained to maximize the likelihood of those images under the model specified by the renderer. The quality of the inferred scene representations produced is visually inspected in Fig.~\ref{3d-single}b. The network accurately and reliably infers the identity and pose of the object present in the scene. In contrast, an identical network trained to predict the ground-truth identity and pose values of the training data (in a similar style to \cite{kulkarni15a}) has much more difficulty in accurately determining the cube's orientation (Fig.~\ref{3d-single}c). The supervised loss forces the network to predict the exact angle of rotation. However this is not identifiable from the image due to rotational symmetry, which leads to conditional probabilities that are multi-modal and difficult to represent using standard network architectures. We also compare with direct optimization of the likelihood from scratch for every test image (Fig.~\ref{3d-single}d), and observe that this method is slower, less stable and more susceptible to local minima. So not only does amortization reduce the cost of inference, but it also overcomes the pitfalls of independent gradient optimization.

\begin{figure}
\begin{center}
\includegraphics[width=0.2\linewidth]{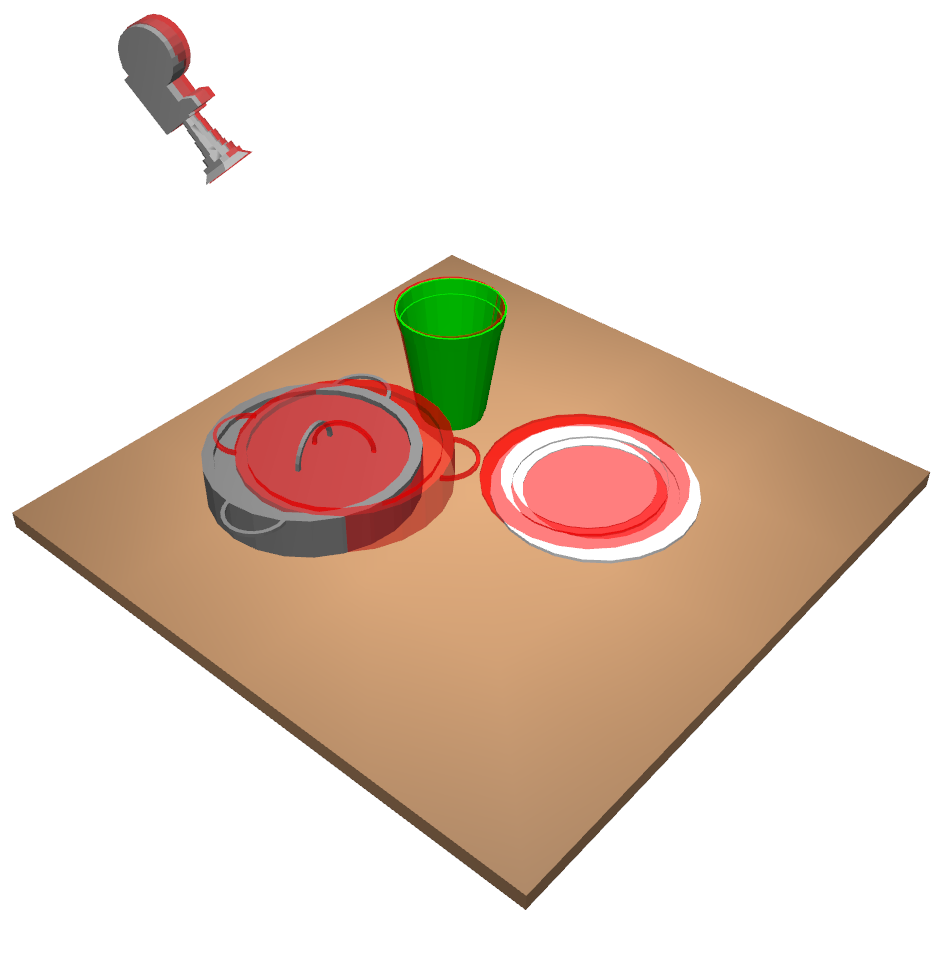}
\hfill
\includegraphics[width=0.25\linewidth]{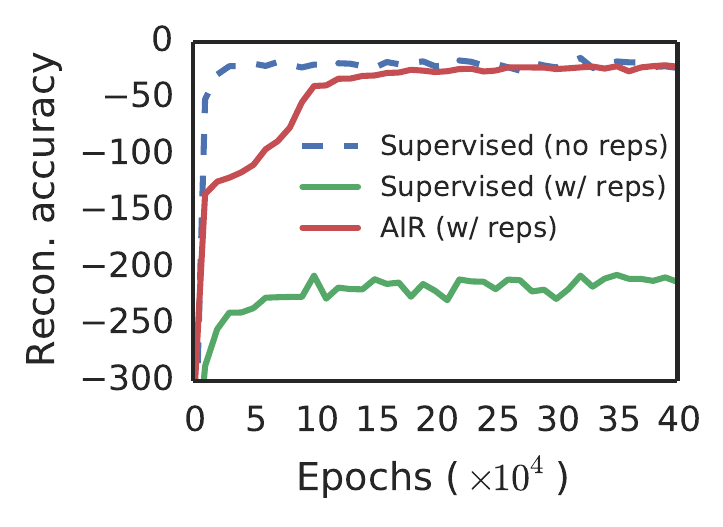}
\hfill
\includegraphics[width=0.24\linewidth]{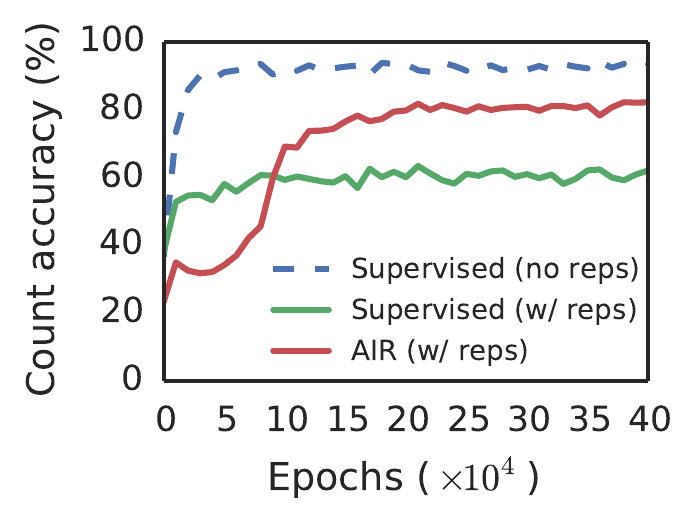}
\hfill
\begin{tikzpicture}
\matrix [matrix of nodes, column sep=0.8mm, row sep=-1mm, every node/.style={inner sep=0, outer sep=0, anchor=center}]
{
  \includegraphics[width=1cm]{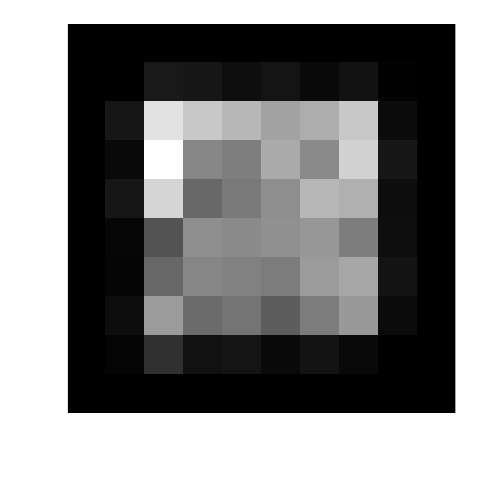} &
  \includegraphics[width=1cm]{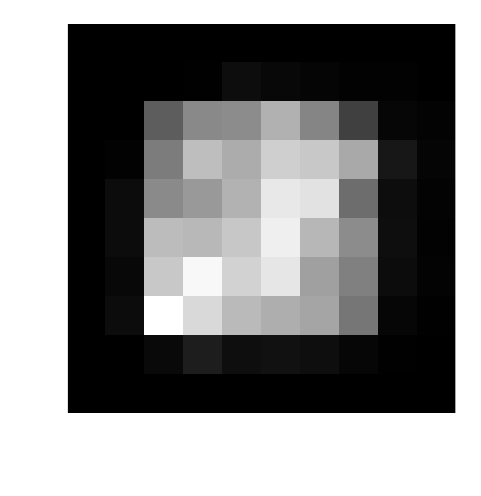} &
  \includegraphics[width=1cm]{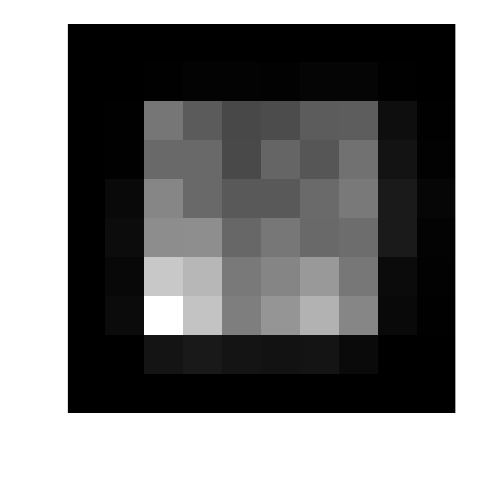} \\
};
\node at (0.26, -1.35) {\includegraphics[width=0.25\linewidth]{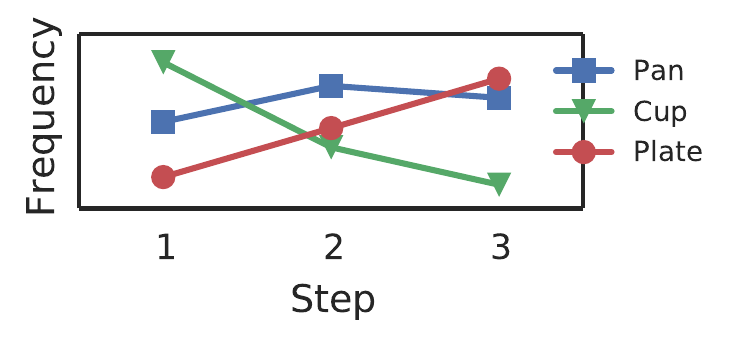}};
\end{tikzpicture}
\caption{\textbf{3D scenes details:} \textit{Left:} Ground-truth object and camera positions with inferred positions overlayed in red (note that inferred cup is closely aligned with ground-truth, thus not clearly visible). We demonstrate fast inference of all relevant scene elements using the \MDL framework. \textit{Middle:} \MDL produces significantly better reconstructions and count accuracies than a supervised method on data that contains repetitions, and is even competitive on simpler data. \textit{Right:} Heatmap of object locations at each time-step (top). The learned policy appears to be more dependent on identity (bottom).}
\label{3d-montage}
\end{center}
\end{figure}

We finally consider a more complex setup, where we infer the counts, identities and positions of a variable number of crockery items, as well as the camera position, in a table-top scene. This would be of critical importance to a robot, say, which is tasked with clearing the table. The goal is to learn to perform this task with as little supervision as possible, and indeed we observe that with \MDL it is possible to do so with no supervision other than a specification of the renderer. We show reconstructions of \MDL's inferences on generated data, as well as real images of a table with varying numbers of plates, in Fig.~\ref{3d-single} and Fig.~\ref{3d-montage}. \MDL's inferences of counts, identities and positions are accurate for the most part. For transfer to real scenes we perform random color and size pertubations to rendered objects during training, however we note that robust transfer remains a challenging problem in general. We provide a quantitative comparison of \MDL's inference robustness and accuracy on generated scenes with that of a fully supervised network in Fig.~\ref{3d-montage}. We consider two scenarios: one where each object type only appears exactly once, and one where objects can repeat in the scene. A naive supervised setup struggles with object repetitions or when an arbitrary ordering of the objects is imposed by the labels, however training is more straightforward when there are no repetitions. \MDL achieves competitive reconstruction and counts despite the added difficulty of object repetitions. 

\section{Related Work}

Deep neural networks have had great success in learning to predict various quantities from images, e.g.,\ object classes \cite{krizhevsky2012}, camera positions \cite{kendall2015} and actions \cite{mnih2015}. These methods work best when large labeled datasets are available for training. At the other end of the spectrum, e.g.,\ in `vision as inverse graphics', only a generative model is specified in advance and prediction is treated as an inference problem, which is then solved using MCMC or message passing at test-time. These models range from highly specified \cite{milch2005,mansinghka2013}, to partially specified \cite{zhu2006,tang13,tang14}, to largely unspecified \cite{salakhutdinov2009}. Inference is very challenging and almost always the bottle-neck in model design. 

Several works exploit data-driven predictions to empower the `vision as inverse graphics' paradigm \cite{hinton1995,jampani2015}. For instance, in PICTURE \cite{kulkarni15a}, the authors use a deep network to distill the results of slow MCMC, speeding up predictions at test-time. Variational auto-encoders \cite{rezende2014,kingma2013} and their discrete counterparts \cite{mnih2014b} made the important contribution of showing how the gradient computations for learning of amortized inference and generative models could be interleaved, allowing both to be learned simultaneously in an end-to-end fashion (see also \cite{schulman2015}). Works like that of \cite{kulkarni2015b} aim to learn disentangled representations in an auto-encoding framework using special network structures and / or careful training schemes. It is also worth noting that attention mechanisms in neural networks have been studied in discriminative and generative settings, e.g.,\ \cite{mnih2014a,jaderberg2015,gregor2015}.

AIR draws upon, extends and links these ideas. By its nature \MDL is also related to the following problems: counting \cite{lempitsky2010,zhang2015}, pondering \cite{graves2016}, and gradient estimation through renderers \cite{loper2014}. It is the combination of these elements that unlocks the full capabilities of the proposed approach.

\section{Discussion}

In this paper our aim has been to learn unsupervised models that are good at scene understanding, in addition to scene reconstruction. We presented several principled models that learn to count, locate, classify and reconstruct the elements of a scene, and do so in a fraction of a second at test-time. The main ingredients are (a) building in meaning using appropriate structure, (b) amortized inference that is attentive, iterative and variable-length, and (c) end-to-end learning. 

We demonstrated that model structure can provide an important inductive bias that gives rise to interpretable representations that are not easily learned otherwise. We also showed that even for sophisticated models or renderers, fast inference is possible. We do not claim to have found an ideal model for all images; many challenges remain, e.g.,\ the difficulty of working with the reconstruction loss and that of designing models rich enough to capture all natural factors of variability. 

Learning in \MDL is most successful when the variance of the gradients is low and the likelihood is well suited to the data. It will be of interest to examine the scaling of variance with the number of objects and alternative likelihoods. It is straightforward to extend the framework to semi- or fully-supervised settings. Furthermore, the framework admits a plug-and-play approach where existing state-of-the-art detectors, classifiers and renderers are used as sub-components of an \MDL inference network. We plan to investigate these lines of research in future work.


\clearpage

{\setstretch{0.0}
\bibliography{bibliography}
\bibliographystyle{plain}
}

\normalsize

\clearpage

\appendix

\section{Stochastic Gradient Estimators}
In this section, we give further details behind the equations in Sec.~\ref{approach}. We simplify notation by not referencing the model parameters $\theta$ and considering a single latent $z$ at a time. Assume we have a function $\ell(z)$ and distribution $q_\phi(z)$; we wish to estimate $\nabla_\phi \mathbb{E}[\ell(z)]$.

\subsection{Reparameterization trick}
\label{appendix:reparameterization}

As per the main body, we supposed the existence of a differentiable function $h$ and random variable $\xi$ with fixed noise distribution $p_\xi(\cdot)$ such that $h(\xi,\phi) \sim q_\phi(\cdot)$. It follows that:
\begin{align}
\partialAt{}{\phi}\mathbb{E}_{z\sim q_\phi}[\ell(z)]  = & \partialAt{}{\phi} \mathbb{E}_{\xi \sim p_\xi} [\ell(h(\xi,\phi))] \notag \\
= & \mathbb{E}_{\xi \sim p_\xi} \ls\partialAt{}{\phi}  \ell(h(\xi,\phi)) \rs\notag\\ 
= & \mathbb{E}_{\xi \sim p_\xi} \ls\partialAt{\ell}{z} \partialAt{h}{\phi}\rs\notag \\
= & \mathbb{E}_{z\sim q_\phi} \ls\partialAt{\ell}{z} \partialAt{h}{\phi}\rs \notag \\
\approx & \partialAt{\ell(z)}{z} \partialAt{h(\xi,\phi)}{\phi}.
\end{align}
In other words, an estimate of the gradient can be recovered by forwarding sampling the model by using the reparameterization given by h, and backpropagating normally through $h$.

\subsection{Likelihood ratio estimator}
\label{appendix:score-function}

The likelihood ratio method simply uses the equality:
\begin{align}
\frac{\partial \log q_\phi(z)}{\partial\phi}  = \frac{\displaystyle \frac{\partial q_\phi(z)}{\partial\phi}}{q_\phi(z)}
\end{align}
to rewrite an integral as an expectation. Assuming that $\frac{\partial q_\phi(z)}{\partial\phi}$ exists and is continuous, we have:
\begin{align}
\frac{\partial}{\partial\phi} \int q_\phi(z) \ell(z) \partial z & = \int_z \frac{\partial q_\phi(z)}{\partial\phi} q_\phi(z) \diffd z \notag \\
& = \int_z \frac{\partial \log \ell_\phi(z)}{\partial\theta} \ell_\phi(z) \ell(z) \diffd z \notag \\
& = \expectationE{\frac{\partial \log q_\phi(z)}{\partial\phi} \ell(z)}{q_\phi(z)} \notag \\
& \approx \frac{\partial \log q_\phi(z)}{\partial\phi} \ell(z).
\end{align}
Note that if $\ell(z)$ is a constant with respect to $z$, then the expression is clearly $0$, since the integral evaluates to the same constant.

\section{Prior for Unary Encoding}

Recall that we can encode the number of objects $n$ as a variable length unary code vector $\zpresvec$ defined by $\zpres^i=1$ for $i\leq n$, and $\zpres^{n+1}=0$ (more generally, it can be useful to implicitly define $\zpres^j=0$, for $j>n$). Consider an arbitrary distribution $p(\cdot)$ over $n$, and denote $\mu_{\geq n}=\sum_{k\geq n} p(k)$ the probability that there are at least $n$ objects. 
We define a joint probability distribution for $\zpresvec$ and show it is consistent with $p(n)$. 

Let $p(\zpres^i=1|\zpres^{i-1})=\zpres^{i-1}\frac{\mu_{\geq i}}{\mu_{\geq (i-1)}}$ for $i\geq 2$, and $p(\zpres^1)=\mu_{\geq 1}$. Note that if $\zpres^i=0$ for any $i$, it follows immediately that $\zpres^j=0$ for $j\geq i$. The sampled vector is therefore a correct unary code. Furthermore,
\begin{align*}
& P(\max\{i: \zpres^i=1\}=n) \\
& = P(\zpres^1=1, \zpres^2=1,\ldots,\zpres^n=1, \zpres^{n+1}=0) \\
& = \left(\prod_{i=1}^n P(\zpres^i=1|\zpres^{i-1}=1)\right) P(\zpres^{n+1}=0|\zpres^n=1) \\
& = \mu_{\geq 1} \times \frac{\mu_{\geq 2}}{\mu_{\geq 1}} \times \frac{\mu_{\geq 3}}{\mu_{\geq 2}} \ldots \frac{\mu_{\geq n}}{\mu_{\geq (n-1)}} \times \left(1-\frac{\mu_{\geq (n+1)}}{\mu_{\geq n}}\right) \\
& = \mu_{\geq n} - \mu_{\geq (n+1)} \\
& = p(n)
\end{align*}
It follows that for $\zpresvec$ following the distribution specified above, the corresponding maximum index is distributed according to $p(n)$ as desired.

\section{Details of 2D Experiments}

All experiments were performed with a batch size of 64. Inference networks and decoders were trained using a learning rate of $10^{-4}$ and baselines were trained using a higher learning rate of $10^{-3}$. LSTMs had 256 cell units and object appearances were coded with 50 units. Images were normalized to hold values between 0 and 1 and the likelihood function was a Gaussian with fixed standard deviation equal to 0.3. The prior $p(n)$ was fixed to a geometric distribution which favors sparse reconstructions.

\section{Details of the \DMDL Network}
\label{appendix:dmdl}

\begin{figure}
\begin{center}
\includegraphics[height=3cm]{misc/model-mdl-mnist-no-bias}
\hfill
\includegraphics[height=3cm]{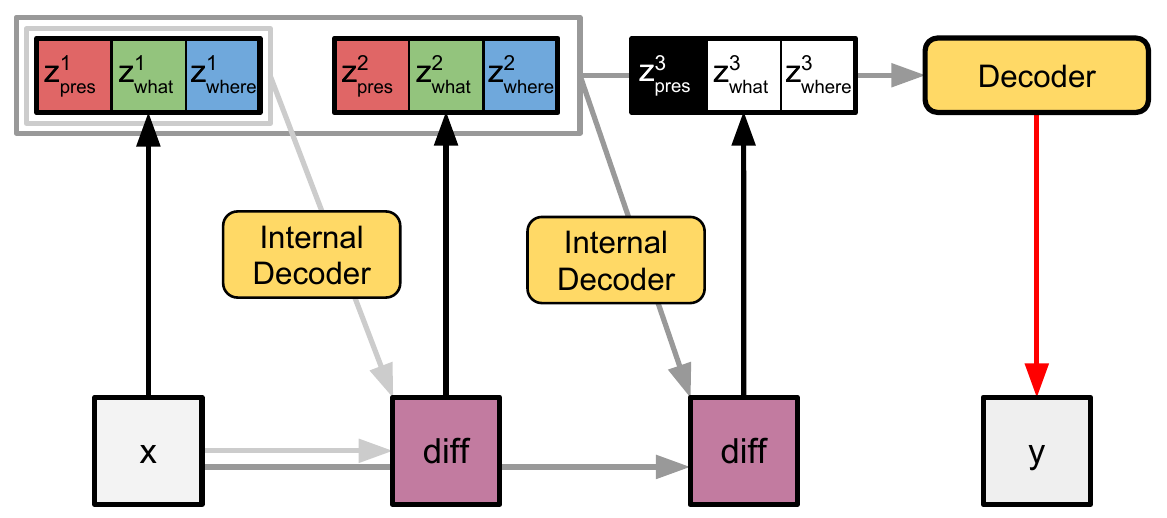}
\caption{\textbf{\MDL vs. \DMDL:} \textit{Left:} The standard \MDL architecture. \textit{Right:} The \DMDL architecture. At each time-step $i$, the latent variables produced so far are used to perform a partial rendering of the scene. The difference of this partial rendering from the image under question is used infer $\zpres^i$, $\zwhat^i$ and $\zwhere^i$ in the current time-step.}
\label{models-dmdl-mnist}
\end{center}
\end{figure}

We assume that the renderer likelihood $p^x(\x|\z^1,\z^2,\ldots,\z^{i})$ has a link function $I$ which maps a sufficient statistic $h^{i}$ to the mean; $h^i$ can be iteratively updated from $h^{i-1}$ and $z^{i-1}$. this is the case for instance for Gaussian and Bernoulli distributions (where $h^i$ is respectively taken to be the mean and log-odds of the distribution). In DAIR, we use the error $\Delta x^i$ between the partial reconstruction $I(h^{i-1})$ and the data $\x$ as inputs to a feed-forward neural network which predicts $\z^i,\zpres^i$. DAIR can be thought of as a special case of AIR with additional structure; namely, the recurrent aspect of AIR is fixed to become a canvas-reconstruction network; see Fig. \ref{models-dmdl-mnist} for more details.

\section{Details of \MDL vs. CNN vs. CAE vs. DRAW Experiments}

The convolutional neural network uses a 64$\times$(5$\times$5)-64$\times$(5$\times$5)-64$\times$(5$\times$5)-512 architecture. 

The convolutional autoencoder uses a sequence of 3 $64\times(6\times 6)$ (for slightly increased performance over $5\times 5$ filters) convolutions with $2\times 2$ max-pooling layers for the encoding, and 3 full convolutions (of the same sizes) and a $2\times 2$ nearest neighbor upsampler for the deconvolution.

The embeddings created by \MDL, DRAW, or CAE are fed through a 4-layer network (each with $512$ units) to produce the 19-way prediction of the sum or a 2-way prediction of the order. 

\section{DRAW Comparisons}
\label{appendix:draw}

We compare \MDL and \DMDL to a state of the art DRAW network with $40$ drawing steps with $4$ latent units per time step, $400$ LSTM hidden units, spatial transformer \cite{jaderberg2015} attention module, and single read and write heads of size $16\times16$. We report free energy on two test sets: a test dataset with $0$, $1$ or $2$ digits, and another with images with precisely $3$ digits. The likelihood model was in all cases Gaussian with fixed standard deviation of $0.3$. 
DRAW outperforms AIR and DAIR on the $0/1/2$ dataset; this is likely due to the fact that DRAW uses many more drawing steps (40) than AIR and thus has an excellent statistical model of single digits. DRAW however does not conceptually understand them as distinct units, as evidenced by its poor generalization on the 3-digits dataset, where \DMDL has both better score, and more meaningful reconstruction: \DMDL partially generalizes to a number of digit never seen (Fig.~\ref{mnist-generalisation-curves}), while DRAW interestingly learns to perfectly ignore exactly one digit in the image (see Fig.~\ref{mnist-generalisation-curves}). More generally, the VAE subroutine present in AIR could be replaced by a DRAW network, thus leading to a `best of both worlds' model with excellent single digit model and understanding of a scene in terms of its constituent parts.

\begin{table}
\begin{center}
\begin{tabular}{ccc} 
\toprule  
\textbf{Model}      & \multicolumn{2}{c}{\textbf{Free Energy}} \\
\cmidrule(r){2-3}
                    & Up to 2 digits    & Only 3 digits \\
\midrule
DRAW                & $\mathbf{-637}$   & $-406$ \\ 
\MDL                & $-620$            & $-316 $ \\ 
\DMDL               & $-611 $           & $\mathbf{-424}$ \\ 
\bottomrule
\end{tabular}
\end{center}
\caption{\textbf{Comparisons with state-of-the-art.} DRAW achieves lower scores than \MDL and \DMDL on up to 2 digits but is outperformed by \DMDL when generalizing to 3 digits.}
\label{draw-ll}
\end{table}

\section{Omniglot Experiments}
\label{appendix:omniglot}

\begin{figure}
\begin{center}
\begin{tikzpicture}
\matrix [matrix of nodes, column sep=1mm, row sep=1mm, every node/.style={inner sep=0, outer sep=0, anchor=center}]
{
  \node[rotate=90]{\small{Data}}; & 
  \includegraphics[height=1cm, width=1cm]{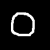} & 
  \includegraphics[height=1cm, width=1cm]{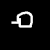} & 
  \includegraphics[height=1cm, width=1cm]{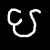} & 
  \includegraphics[height=1cm, width=1cm]{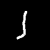} & 
  \includegraphics[height=1cm, width=1cm]{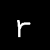} & 
  \includegraphics[height=1cm, width=1cm]{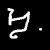} & 
  \includegraphics[height=1cm, width=1cm]{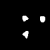} \\
  \vspace{1mm} \\
  \node[rotate=90]{\small{Step 1}}; & 
  \includegraphics[height=1cm, width=1cm]{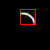} &
  \includegraphics[height=1cm, width=1cm]{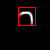} & 
  \includegraphics[height=1cm, width=1cm]{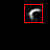} & 
  \includegraphics[height=1cm, width=1cm]{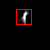} & 
  \includegraphics[height=1cm, width=1cm]{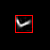} & 
  \includegraphics[height=1cm, width=1cm]{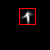} & 
  \includegraphics[height=1cm, width=1cm]{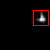} \\
  \node[rotate=90]{\small{Step 2}}; & 
  \includegraphics[height=1cm, width=1cm]{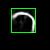} &
  \includegraphics[height=1cm, width=1cm]{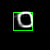} & 
  \includegraphics[height=1cm, width=1cm]{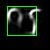} & 
  \includegraphics[height=1cm, width=1cm]{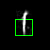} & 
  \includegraphics[height=1cm, width=1cm]{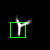} & 
  \includegraphics[height=1cm, width=1cm]{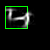} & 
  \includegraphics[height=1cm, width=1cm]{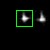} \\
  \node[rotate=90]{\small{Step 3}}; & 
  \includegraphics[height=1cm, width=1cm]{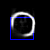} & 
  \includegraphics[height=1cm, width=1cm]{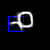} & 
  \includegraphics[height=1cm, width=1cm]{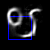} & 
  \includegraphics[height=1cm, width=1cm]{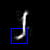} & 
  \includegraphics[height=1cm, width=1cm]{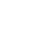} & 
  \includegraphics[height=1cm, width=1cm]{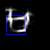} & 
  \includegraphics[height=1cm, width=1cm]{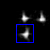} \\
  \node[rotate=90]{\small{Step 4}}; & 
  \includegraphics[height=1cm, width=1cm]{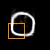} & 
  \includegraphics[height=1cm, width=1cm]{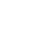} & 
  \includegraphics[height=1cm, width=1cm]{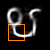} & 
  \includegraphics[height=1cm, width=1cm]{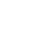} & 
  \includegraphics[height=1cm, width=1cm]{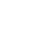} & 
  \includegraphics[height=1cm, width=1cm]{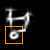} & 
  \includegraphics[height=1cm, width=1cm]{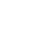} \\
};
\end{tikzpicture}
\caption{\textbf{Omniglot:} \MDL reconstructions at every time-step. \MDL uses variable numbers of strokes for digits of varying complexity.}
\label{omniglot-reconstructions-over-time}
\end{center}
\end{figure}

We also investigate the behavior of \MDL on the Omniglot dataset \cite{lake2015} which contains 1623 different handwritten characters from 50 different alphabets. Each of the 1623 characters was drawn online via Amazon's Mechanical Turk by 20 people. This means that the data was produced according a process (pen strokes) that is not directly reflected in the structure of our generative model. It is therefore interesting to examine the outcome of learning under mis-specification. We train the model from the previous section, this time allowing for a maximum of up to 4 inference time-steps per image. Fig.~\ref{omniglot-reconstructions-over-time} shows that by using different numbers of time-steps to describe characters of varying complexity, \MDL discovers a representation consisting of spatially coherent elements resembling strokes, despite not exploiting stroke labels in the data or building in the physics of strokes, in contrast with \cite{lake2015}. Further results can be found in the supplementary video.

\section{Sprites Experiments}
\label{appendix:sprites}

\begin{figure}
\begin{center}
\begin{tikzpicture}
\matrix [matrix of nodes, column sep=1mm, row sep=1mm, every node/.style={inner sep=0, outer sep=0, anchor=center}]
{
  \small{\textit{(a)}} & 
  \includegraphics[height=1cm, width=1cm]{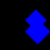} & 
  \includegraphics[height=1cm, width=1cm]{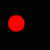} & 
  \includegraphics[height=1cm, width=1cm]{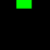} & 
  \includegraphics[height=1cm, width=1cm]{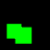} & 
  \includegraphics[height=1cm, width=1cm]{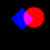} & 
  \includegraphics[height=1cm, width=1cm]{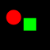} & 
  \includegraphics[height=1cm, width=1cm]{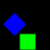} \\
  \small{\textit{(b)}} & 
  \includegraphics[height=1cm, width=1cm]{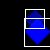} & 
  \includegraphics[height=1cm, width=1cm]{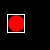} & 
  \includegraphics[height=1cm, width=1cm]{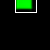} & 
  \includegraphics[height=1cm, width=1cm]{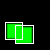} & 
  \includegraphics[height=1cm, width=1cm]{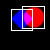} & 
  \includegraphics[height=1cm, width=1cm]{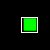} & 
  \includegraphics[height=1cm, width=1cm]{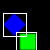} \\
  \small{\textit{(c)}} & 
  \includegraphics[height=1cm, width=1cm]{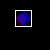} & 
  \includegraphics[height=1cm, width=1cm]{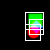} & 
  \includegraphics[height=1cm, width=1cm]{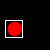} & 
  \includegraphics[height=1cm, width=1cm]{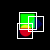} & 
  \includegraphics[height=1cm, width=1cm]{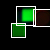} & 
  \includegraphics[height=1cm, width=1cm]{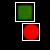} & 
  \includegraphics[height=1cm, width=1cm]{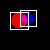} \\
};
\end{tikzpicture}
\caption{\textbf{Sprites overview:} (a) Images from the dataset. (b) \MDL reconstructions. We visualize the model's attention at every time-step (inferred object boundaries) in white. (c) A selection of samples from the learned model.}
\label{sprites-data-reconstructions-samples}
\end{center}
\end{figure}

We also consider a 50$\times$50 dataset of sprites: red circles, green squares and blue diamonds. Each image in the dataset contains zero, one or two sprites (see Fig.~\ref{sprites-data-reconstructions-samples}a). The images are composed additively (sprites do not occlude each other). We use the exact same model structure as for the multi-MNIST dataset.

\begin{figure}
\begin{center}
\includegraphics[width=0.29\linewidth]{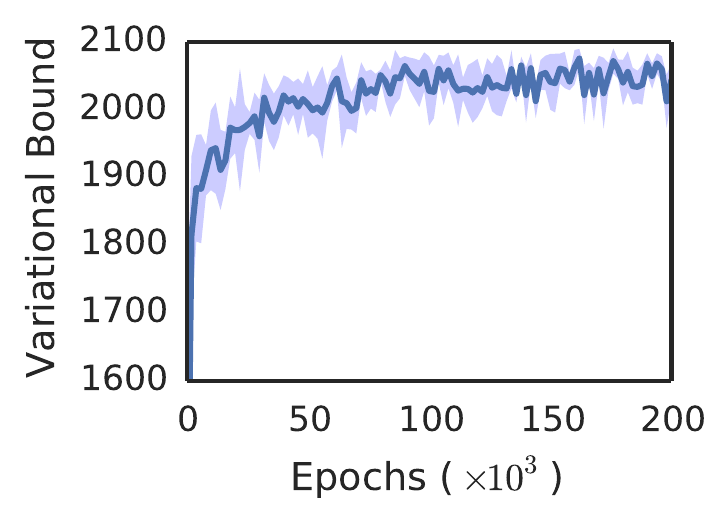}
\includegraphics[width=0.29\linewidth]{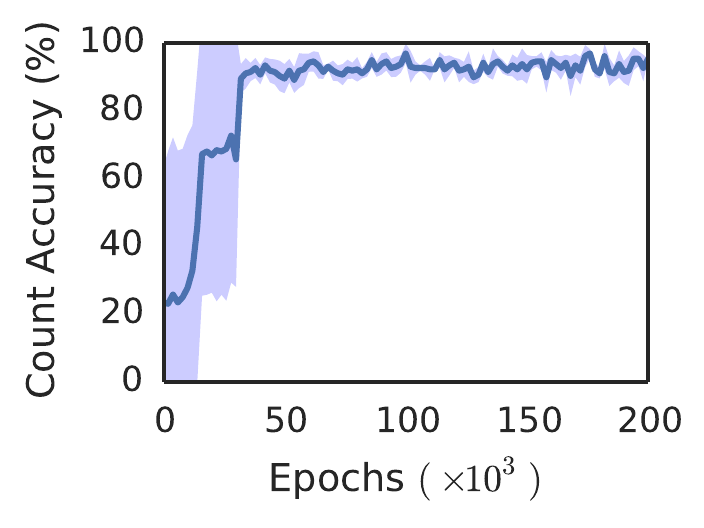}
\caption{\textbf{Sprites quantitative results:} \textit{Left:} Variational lower bound over the course of training. \textit{Right:} Sprite count accuracy.}
\label{sprites-curves}
\end{center}
\end{figure}

\begin{figure}
\begin{center}
\begin{tikzpicture}
\matrix [matrix of nodes, column sep=1mm, row sep=1mm, every node/.style={inner sep=0, outer sep=0, anchor=center}]
{
  \node[rotate=90]{\small{Data}}; & 
  \includegraphics[height=1cm, width=1cm]{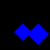} & 
  \includegraphics[height=1cm, width=1cm]{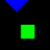} & 
  \includegraphics[height=1cm, width=1cm]{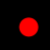} & 
  \includegraphics[height=1cm, width=1cm]{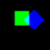} & 
  \includegraphics[height=1cm, width=1cm]{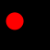} & 
  \includegraphics[height=1cm, width=1cm]{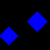} & 
  \includegraphics[height=1cm, width=1cm]{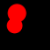} \\
  \vspace{1mm} \\
  \node[rotate=90]{\small{0}}; & 
  \includegraphics[height=1cm, width=1cm]{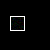} & 
  \includegraphics[height=1cm, width=1cm]{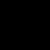} & 
  \includegraphics[height=1cm, width=1cm]{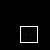} & 
  \includegraphics[height=1cm, width=1cm]{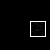} & 
  \includegraphics[height=1cm, width=1cm]{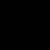} & 
  \includegraphics[height=1cm, width=1cm]{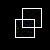} & 
  \includegraphics[height=1cm, width=1cm]{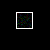} \\
  \node[rotate=90]{\small{1k}}; & 
  \includegraphics[height=1cm, width=1cm]{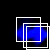} & 
  \includegraphics[height=1cm, width=1cm]{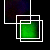} & 
  \includegraphics[height=1cm, width=1cm]{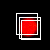} & 
  \includegraphics[height=1cm, width=1cm]{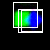} & 
  \includegraphics[height=1cm, width=1cm]{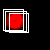} & 
  \includegraphics[height=1cm, width=1cm]{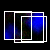} & 
  \includegraphics[height=1cm, width=1cm]{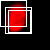} \\
  \node[rotate=90]{\small{10k}}; & 
  \includegraphics[height=1cm, width=1cm]{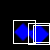} & 
  \includegraphics[height=1cm, width=1cm]{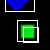} & 
  \includegraphics[height=1cm, width=1cm]{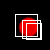} & 
  \includegraphics[height=1cm, width=1cm]{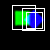} & 
  \includegraphics[height=1cm, width=1cm]{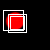} & 
  \includegraphics[height=1cm, width=1cm]{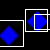} & 
  \includegraphics[height=1cm, width=1cm]{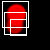} \\
  \node[rotate=90]{\small{200k}}; &  
  \includegraphics[height=1cm, width=1cm]{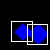} & 
  \includegraphics[height=1cm, width=1cm]{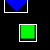} & 
  \includegraphics[height=1cm, width=1cm]{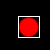} & 
  \includegraphics[height=1cm, width=1cm]{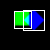} & 
  \includegraphics[height=1cm, width=1cm]{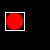} & 
  \includegraphics[height=1cm, width=1cm]{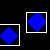} & 
  \includegraphics[height=1cm, width=1cm]{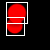} \\
};
\end{tikzpicture}
\caption{\textbf{Sprites learning:} \textit{Top:} Images from the dataset. \textit{Bottom:} Reconstructions at different points during training. A video of this sequence is included in the supplementary material.}
\label{sprites-reconstructions-over-time}
\end{center}
\end{figure}

\begin{figure}[t!]
\begin{center}
\begin{tikzpicture}
\matrix [matrix of nodes, column sep=3mm, row sep=1mm, every node/.style={inner sep=0, outer sep=0, anchor=center}]
{
  \draw[thick,-stealth] (-0.2,-0.2) -- (0.2,0.2); & \includegraphics[width=1.3cm]{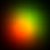} & &
  \draw[thick,-stealth] (0.2,-0.2) -- (-0.2,0.2); & \includegraphics[width=1.3cm]{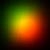} & &
  \draw[thick,-stealth] (0.2,0.2) -- (-0.2,-0.2); & \includegraphics[width=1.3cm]{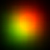} \\
};
\end{tikzpicture}
\caption{\textbf{Sprites scanning policies:} Empirical heatmaps of where the attention windows go to (red, and green for the first and second time-steps respectively). As expected, the policy is random. Each figure is for a different inference network that has been trained from scratch using a different seed. This suggests that the policy is spatial, as opposed to identity- or size-based.}
\label{sprites-visitation-maps}
\end{center}
\end{figure}

At the end of unsupervised training, \MDL successfully learns about the underlying causes of the scenes (namely, the sprites), as well as their counts and locations, and also produces convincing reconstructions (see Fig.~\ref{sprites-data-reconstructions-samples}b). Note that the inference network correctly detects the correct number of sprites even when two overlapping sprites of the same type and color appear in the same image (Fig.~\ref{sprites-data-reconstructions-samples}a,b, images 1 and 3). Also note that the reconstructions are accurate, meaning that the inference network successfully produces the codes for each sprite despite the presence of the other sprites in its field of view. Fig.~\ref{sprites-data-reconstructions-samples}c displays a collection of samples from the model after training. We display quantitative evaluation of the network's counting accuracy in Fig.~\ref{sprites-curves}, reconstructions over the course of learning in Fig.~\ref{sprites-reconstructions-over-time}, and a visualization of its scanning policy in Fig.~\ref{sprites-visitation-maps}.

Note that these tasks can only be successfully achieved once the inference network has learned a sensible policy for scanning the image, e.g.,\ one in which every object is attended to only once. However the network must break multiple symmetries to achieve this, e.g.,\ it does not matter which object it explains first. In Fig.~\ref{sprites-visitation-maps} we visualize the learned scanning policy for 3 different runs of training (only differing in the random seed). In each case a unique policy is learned, and the policy appears to be spatial (as opposed to one that is based on digit identity or size).

\section{Details of 3D Scene Experiments}
\label{appendix:3d}

The experiments in section \ref{3d} were performed using the rendering capabilities of the MuJoCo physics simulator \cite{todorov2012}. 

\subsection{Gradient estimation}

Differentiation of MuJoCo's graphics engine was performed using forward finite-differencing (with a constant $\epsilon=10^{-4}$) with respect to the scene configuration. This is a generic procedure which would work for any graphics engine; we chose MuJoCo because it is fast (using only the fixed functionality of OpenGL) and because scenes are conveniently parameterized. Interestingly, despite the coarse 8-bit output of OpenGL, quantization errors appeared to average out reasonably well over the pixels.

\subsection{Scene generation}

\begin{figure}
\begin{center}
\begin{tikzpicture}
\matrix [matrix of nodes, column sep=1mm, row sep=1mm, every node/.style={inner sep=0, outer sep=0, anchor=center}]
{
  \includegraphics[height=4.5cm]{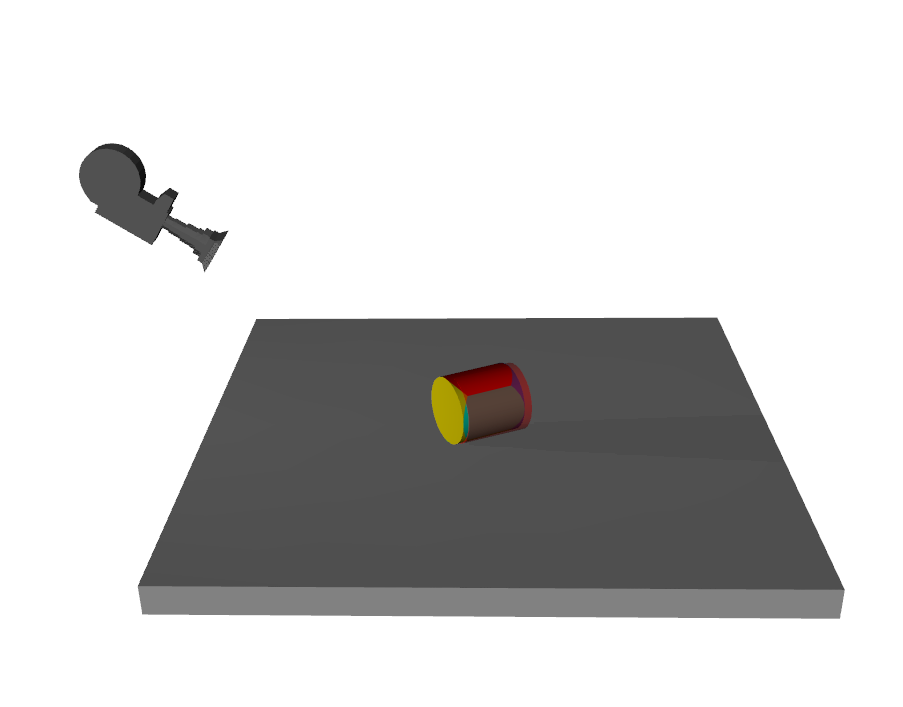} & 
  \includegraphics[height=2cm]{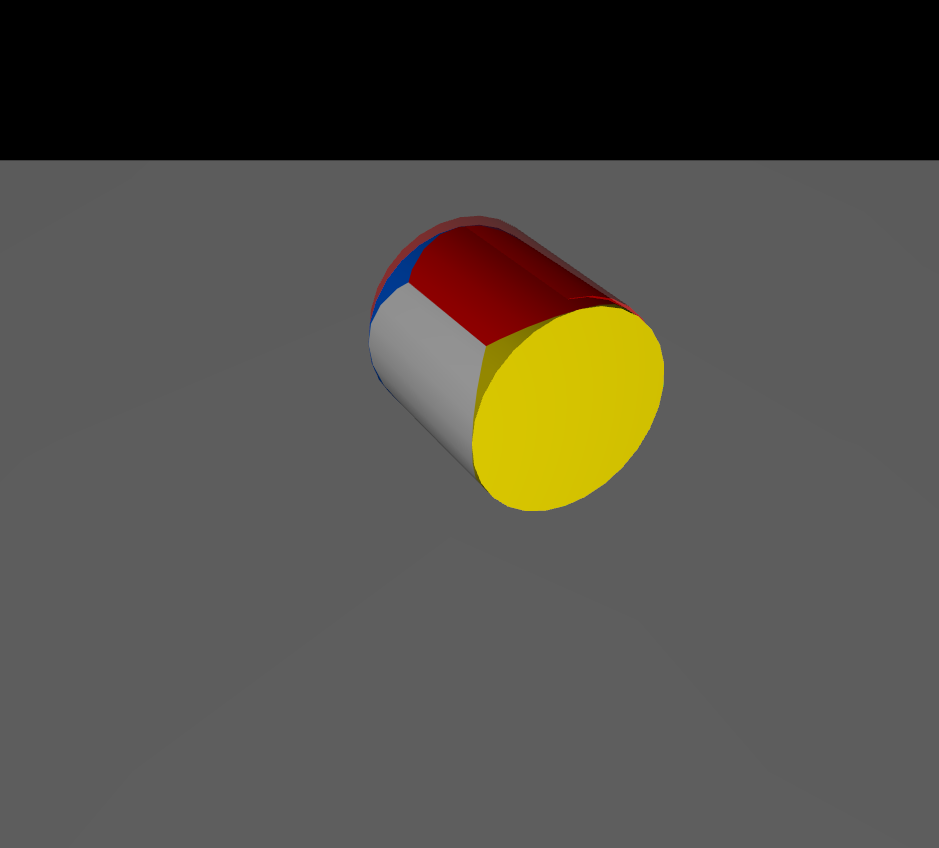} \\
};
\end{tikzpicture}
\caption{\textit{Left}: Illustration of the setup for single-object scenes similar to Fig.~\ref{3d-montage} in the main text. The illustration shows the fixed camera, the ground truth object (textured cylinder), and an example inference (transparent red). \textit{Right}: Rendering from camera as fed into the inference network (before downsampling).}
\label{3d-single-montage}
\end{center}
\end{figure}

\paragraph{Single object scenes:} For the results shown in Fig.~\ref{3d-single} we created a scene that contained a MuJoCo box geom representing the table, 3 `objects' (also in the form of MuJoCo geoms; cube, sphere, textured cylinder), and a fixed camera. The objects could be moved in the plane of the table and rotated along the axis orthogonal to it (i.e.\ 3 degrees of freedom per object). We created random scenes containing at most one object by randomly sampling position, rotation angle, object presence (visibility) and object type. (Geoms were made invisible by moving them out of the field of view of the camera.) An illustration is shown in Fig.~\ref{3d-single-montage}.

\paragraph{Tabletop scenes:}
For the results shown in Fig.~\ref{3d-single} and \ref{3d-montage} we used scenes with a box geom for the table, and nine mesh geoms for the crockery items. The cup, pan, and plate were each replicated three times to allow for arbitrary three-objects scenes. Each geom had three degrees of freedom (position in the table plane and rotation). Random scenes with up to $N=3$ objects were created by uniformly sampling position, rotation angle, object presence, and object type three times. As for the single objects were rendered invisible by moving them outside of the field of view of the camera.

\begin{figure}
\begin{center}
\begin{tikzpicture}
\matrix [matrix of nodes, column sep=3mm, row sep=3mm, every node/.style={inner sep=0, outer sep=0, anchor=center}]
{
  \node[rotate=90]{\small{\textit{(a)} Data}}; &
  \includegraphics[height=2.4cm, width=2.4cm]{3d/table-camera/trueSceneA_1_3} & 
  \includegraphics[height=2.4cm, width=2.4cm]{3d/table-camera/trueSceneA_1_7} & 
  \includegraphics[height=2.4cm, width=2.4cm]{3d/table-camera/trueSceneA_3_1} & 
  \includegraphics[height=2.4cm, width=2.4cm]{3d/table-camera/trueSceneA_3_2} & 
  \includegraphics[height=2.4cm, width=2.4cm]{3d/table-camera/trueSceneA_4_9} \\
  \node[rotate=90]{\small{\textit{(b)} Reconstruction}}; & 
  \includegraphics[height=2.4cm, width=2.4cm]{3d/table-camera/samples_1_3_1} & 
  \includegraphics[height=2.4cm, width=2.4cm]{3d/table-camera/samples_1_7_7} & 
  \includegraphics[height=2.4cm, width=2.4cm]{3d/table-camera/samples_3_1_1} & 
  \includegraphics[height=2.4cm, width=2.4cm]{3d/table-camera/samples_3_2_1} & 
  \includegraphics[height=2.4cm, width=2.4cm]{3d/table-camera/samples_4_9_1} \\
  \\
  \node[rotate=90]{\small{\textit{(a)} Data}}; &
  \includegraphics[height=2.4cm, width=2.4cm]{3d/table-camera/trueSceneA_6_6} & 
  \includegraphics[height=2.4cm, width=2.4cm]{3d/table-camera/trueSceneA_9_8} & 
  \includegraphics[height=2.4cm, width=2.4cm]{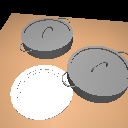} & 
  \includegraphics[height=2.4cm, width=2.4cm]{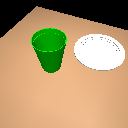} & 
  \includegraphics[height=2.4cm, width=2.4cm]{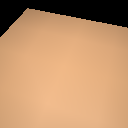} \\ 
  \node[rotate=90]{\small{\textit{(b)} Reconstruction}}; &  
  \includegraphics[height=2.4cm, width=2.4cm]{3d/table-camera/samples_6_6_1} & 
  \includegraphics[height=2.4cm, width=2.4cm]{3d/table-camera/samples_9_8_1} & 
  \includegraphics[height=2.4cm, width=2.4cm]{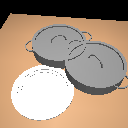} & 
  \includegraphics[height=2.4cm, width=2.4cm]{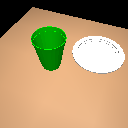} & 
  \includegraphics[height=2.4cm, width=2.4cm]{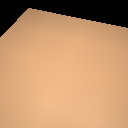}\\
};
\end{tikzpicture}
\caption{\textbf{3D scenes with variable camera:} \MDL results for inferring the camera angle of the scene, as well as the counts, identities and poses of multiple objects in a 3D table-top scene similar to the results presented in Section \ref{3d} in the main text but with the additional complication of an unknown camera angle. (a) Images from the dataset. (b) Reconstruction of the scene description inferred by our AIR network. Note that due to the down-sampling of the images that were used as input to the inference network and the blurring in the likelihood computation accurate estimation of the rotation angle is essentially impossible. }
\label{3d-table-camera}
\end{center}
\end{figure}

We experimented with two versions of the scene: one with a fixed camera, and one version where the camera could be moved in an orbit around the table (i.e.\ one degree of freedom). We discuss the experiment with the fixed camera in the main text. For the latter set of scenes, the camera position was also chosen randomly and the image was rendered from the random camera position. Camera movement was restricted to $\pm$ 40 degrees from the central position. In this experiment the model had to learn to infer the camera position in addition to the objects on the table. The montage in Fig.~\ref{3d-montage} in the main text shows a ground truth scene (with camera) and the inferred identities and positions of the objects as well as the inferred position of the camera. We show several examples of random scenes with variable camera and the associated inferences in Fig.~\ref{3d-table-camera}. For the most part the network infers all scene parameters reliably.

\paragraph{Image preprocessing:} We rendered all scene images at 128 $\times$ 128 pixels. We down-sampled scene images to 32 $\times$ 32 pixels for input to the network.

\subsection{Model}
We trained a network to perform inference in the following fixed generative model:
\begin{align}
& p(\x, \zpresNN, \zwhereNN, \zwhatNN) = \\
& p(\x | \zpresNN, \zwhereNN, \zwhatNN) \prod_{i=1}^N p(\zpres^i)p(\zwhat^i)p(\zwhere^i), \nonumber
\end{align}
where $\zpres^i$ is the visibility indicator: $\zpres^i \sim \textrm{Bernoulli}(\alpha)$ for object $i$; $\zwhere \in \mathbb{R}^3$ indicates position and rotation angle: $\zwhere^i \sim \mathcal{N}(0, \Sigma_\textrm{where})$; and $\zwhat^i$ is a three-valued discrete variable indicating the object type (mesh / geom type): $\zwhat^i \sim \textrm{Discrete}(\beta)$.

The marginal distribution over scenes under this model is the same as the marginal distribution under a model of form described in Section \ref{approach} in the main text where $p(n) = \textrm{Binomial}(N,\alpha)$ and $n = \sum_{i=1}^N z_i$.

For the variable camera scenes the model included an additional random variable $z_\textrm{cam} \in \mathbb{R}$ where $z_\textrm{cam} \sim \mathcal{N}(0, \sigma_\textrm{cam}^2)$.

To evaluate the likelihood term $p(\x|\z)$ we (1) render the scene description using the MuJoCo rendering engine to produce a high-resolution image $\y$; (2) blur the resulting image $\y$ as well as $\x$ using a fixed-with blur kernel; (3) compute $\mathcal{N}(\x | \y, \mathbf{I}\sigma^2_{x})$.

\subsection{Network}
The AIR inference network for our experiments is a standard recurrent network (no LSTM) that is run for a fixed number of steps ($N = 1$ or $N=3$). In each step the network computes:
\begin{equation}
(\omega^i_{\textrm{pres}}, \omega^i_{\textrm{what}}, \omega^i_{\textrm{where}}, \h^i) = R(\x, \zpres^{i-1}, \zwhat^{i-1}, \zwhere^{i-1}, \h^{i-1}), \nonumber
\end{equation}
where the $\omega^i$ represent the parameters of the sampling distributions for the random variables: Bernoulli for $\zpres$; Discrete for $\zwhat$; and Gaussian for $\zwhere$. For the experiments with random camera angle we use a separate network that computes $\omega_\textrm{cam} = F(\x)$ and we provide the sampled camera angle as additional input to $R$ at each time step.

\subsection{Supervised learning}

For the baselines trained in a supervised manner we use the ground truth scene variables $\zpresNN, \zwhereNN, \zwhatNN$ that underly the training scene images as labels and train a network of the same form as the inference network to maximize the conditional log likelihood of the ground truth scene variables given the image. 

\section{Inference Speed}
\label{appendix:speed}

For the MNIST experiments, upon completion of training each inference step takes 5.6 milliseconds on average to execute on an Nvidia Quadro K4000 GPU (a step corresponds to inference of state for a single object), in other words up to around 17 milliseconds per image for images of 3 digits. Therefore running at  around 59 frames per second, inference is significantly faster than real-time.

For 3D scenes, the equivalent numbers are  around 2.3 milliseconds per step and 8 milliseconds per image (due to absence of spatial transformers) on a CPU. Gradient-based optimization is slower, taking 5 milliseconds per gradient step per object, and tens or hundreds of steps per image, depending on the choice of optimizer. 

Training for the MNIST model converges in around 2 days, and in around 3 days for the 3D scenes.

\end{document}